\documentclass[journal]{IEEEtran}
\usepackage{amsmath,amsfonts}
\usepackage{algorithmic}
\usepackage{subfigure}
\usepackage{array}
\usepackage[caption=false,font=footnotesize,labelfont=rm,textfont=rm]{subfig}
\usepackage{textcomp}
\usepackage{stfloats}
\usepackage{url}
\usepackage{verbatim}
\usepackage{graphicx}
\usepackage{cite}
\usepackage{booktabs}
\usepackage{diagbox}
\usepackage{color}
\usepackage{xpatch}
\usepackage{multirow}
\usepackage{multicol}

\usepackage{makecell}
\usepackage{array}
\usepackage{colortbl}
\usepackage{url}
\definecolor{color}{RGB}{0,0,0} 
\definecolor{temp}{RGB}{0,0,0} %转成黑色
\definecolor{blue}{RGB}{0,0,0} %转成黑色
\usepackage{setspace}
\usepackage[ruled,linesnumbered]{algorithm2e} %带行号
\usepackage{tikz}

\newcommand\submittedtext{%
  \footnotesize This work has been submitted to the IEEE for possible publication. Copyright may be transferred without notice, after which this version may no longer be accessible.}
\newcommand\submittednotice{%
\begin{tikzpicture}[remember picture,overlay]
\node[anchor=south,yshift=10pt] at (current page.south) {\fbox{\parbox{\dimexpr0.65\textwidth-\fboxsep-\fboxrule\relax}{\submittedtext}}};
\end{tikzpicture}%
}

\begin{document}
\bstctlcite{IEEEexample:BSTcontrol}

\title{Location-guided Head Pose Estimation for Fisheye Image}

\author{Bing~Li,~Dong~Zhang,~Cheng~Huang,~Yun~Xian,~Ming~Li,~\IEEEmembership{Senior Member,~IEEE,}~and~Dah-Jye~Lee,~\IEEEmembership{Senior Member,~IEEE}
    %\thanks{This work was supported by National Natural Science Foundation of China (62173353), Guangzhou Municipal People's Livelihood Science and Technology Plan (201903010040), Science and Technology Program of Guangzhou, China (202007030011).}
\thanks{Bing~Li,~Dong~Zhang,~Cheng~Huang,~and~Yun~Xian are with the School of Electronics and Information Technology, Sun Yat-sen University, China (e-mail: libing29@mail2.sysu.edu.cn; zhangd@mail.sysu.edu.cn; huangch63@mail2.sysu.edu.cn.;xiany7@mail2.sysu.edu.cn).}
\thanks{Ming~Li is with the Data Science Research Center, Duke Kunshan University, Kunshan, China, 215316. (e-mail:ming.li369@dukekunshan.edu.cn).}        
\thanks{Dah-Jye~Lee is with the Department of Electric and Computer Engineering, Brigham Young University, U.S.A.(e-mail: djlee@byu.edu).}
\thanks{Corresponding author: Dong Zhang (e-mail:zhangd@mail.sysu.edu.cn).}}
\maketitle
\submittednotice
	
\begin{abstract}
Camera with a fisheye or ultra-wide lens covers a wide field of view that cannot be modeled by the perspective projection. Serious fisheye \textcolor{black}{lens} distortion in the peripheral region of the image leads to degraded performance of the \textcolor{black}{existing} head pose estimation models trained on undistorted images. This paper presents a new approach for head pose estimation that uses the knowledge of head location in the image to reduce the negative effect of fisheye distortion. We develop an end-to-end convolutional neural network to estimate the head pose with the multi-task learning of head pose and head location. Our proposed network estimates the head pose directly from the fisheye image without the operation of rectification or calibration. We also created \textcolor{black}{a} fisheye-\textcolor{black}{distorted} version of the three popular head pose estimation datasets, BIWI, 300W-LP, and AFLW2000 for our experiments. Experiments results show that our network remarkably improves the accuracy of head pose estimation compared with other state-of-the-art one-stage and two-stage methods.	
\end{abstract}
	
\begin{IEEEkeywords}
Head pose estimation, fisheye camera, fisheye distortion, convolutional neural networks 
\end{IEEEkeywords}
	
\section{Introduction}
\IEEEPARstart{T}{he} goal of head pose estimation (HPE) in the context of computer vision is to estimate the \textcolor{temp}{orientation} of the head concerning the camera coordinate system. The estimated head pose is usually expressed by Euler angles (pitch, yaw, roll)~\cite{khalilkhan_2021_0}. \textcolor{temp}{Head pose is regarded as an important non-verbal characteristic of human intent for human social behavior analysis~\cite{chih-weichen_2011_6}. It is involved in many cognitive processes and provides strong cues for human attention, motivation, and intention. Head pose estimation has attracted significant research interest in the fields of cognitive science and computer vision as it helps the computer or robot understand human intention and plays an essential role in various intelligent applications such as driver assistance systems~\cite{nawalalioua_2016_40}, security surveillance~\cite{benfold2009guiding}, virtual reality~\cite{george2020gazeroomlock}, human-robot interaction~\cite{wang2018human} and cognitive system~\cite{odobez2007cognitive}.} 

% driver assistance systems~\cite{nawalalioua_2016_40,boon-giinlee_2012_41,murphy2010head}
%\textcolor{black}{Since} accurate estimation of head pose is essential to many applications including human-computer interaction, human social behavior analysis, surveillance, virtual reality, and psychological assessment, head pose estimation has attracted significant research interest in recent years~\cite{nawalalioua_2016_40, kaidicao_2018_42, chih-weichen_2011_6, boon-giinlee_2012_41, yusukesugano_2014_43}. 

Traditionally, research work on HPE was committed to estimating the head pose from rectilinear (conventional) images. With the rapid growth of the fisheye camera market, estimating the head pose from fisheye image is becoming an emerging requirement. 

\begin{figure}[tbp]
    \centering
    \subfigure[]{%
        \includegraphics[height=1.5in]{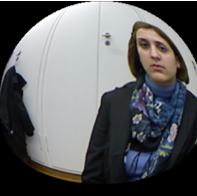}
        %\label{fig:subfigure1}
    }
    \hspace{0.01cm} %插入水平间隔
    \subfigure[]{%
        \includegraphics[height=1.5in]{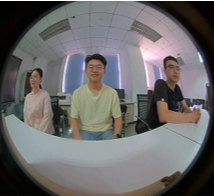}
        %\label{fig:subfigure2}
    }
    \caption{Examples of fisheye images obtained from (a) BIWI-360 dataset, \\ \textcolor{black}{(b) a fisheye camera (JR$^\circledR$HF900).}}
    \label{fig:figure1}
\end{figure}

Rather than generating images with straight lines of perspective, fisheye camera use a special mapping to produce images with a characteristic convex non-rectilinear appearance, as shown in Fig.~\ref{fig:figure1}. This \textcolor{black}{distortion} nature of the fisheye camera makes estimating head pose from fisheye image a very challenging task.	

\textcolor{black}{Head pose estimation} from rectilinear image has already achieved promising results. An intuitive idea \textcolor{black}{for estimating} the head pose from fisheye image is \textcolor{black}{the} so-called two-stage strategy. \textcolor{black}{It involves} rectifying the fisheye image to a rectilinear image first, and then using a good rectilinear-image-based HPE method to estimate the head pose from the rectified image. The \textcolor{black}{main} advantage of this \textcolor{black}{strategy} is that head pose estimators already developed for the rectilinear image can be directly migrated to rectified fisheye image. \textcolor{black}{If the camera parameters are available, this strategy may achieve good results. However, the camera parameters are not always available in real-world scenarios and the camera calibration is proved to be a tedious and time-consuming process. The lack of the camera parameters prevents the rectification operation from completely eliminating the geometrical difference in appearance between the rectified image and the corresponding rectilinear image, which decreases the accuracy of the two-stage based strategy in head pose estimation. 
}

Another approach is to perform head pose estimation directly on the fisheye images, which we call \textcolor{black}{the} one-stage method. One-stage methods do not require an explicit process of image rectification and are expected to result in a unified optimized framework. \textcolor{black}{In image processing, there have been successful attempts to extend traditional techniques designed for rectilinear images to fisheye images or spherical (omnidirectional) images. Hansen et al. mapped a wide-angle image to a sphere and developed a scale invariant feature transform (SIFT) algorithm on the sphere to match points between wide-angle images~\cite{hansen2007scale}. Additionally, Cruz-Mota et al. proposed a SIFT-based algorithm in spherical coordinates for point matching between spherical and rectilinear images~\cite{cruz2012scale}. Demonceaux et al. used the geodesic distance metric between pixels in fisheye image to redefine the basic filter kernels~\cite{demonceaux2011central}. Delibasis proposed an efficient implementation to redefine the Laplacian of Gaussian kernel using the geodesic distance~\cite{delibasis2018efficient}. Georgakopoulos et al. presented a method of human pose recognition for fisheye images by using Convolutional Neural Networks (CNN) enhanced with geodesically corrected Zernike moments~\cite{georgakopoulos2018pose}. These works employed a one-stage approach without resorting to image rectification and provided us with valuable insights and confidence in the practicability of one-stage approaches for head pose estimation.}

In this paper, we present an end-to-end estimation network to predict the Euler angles of head pose from a single fisheye image. Objects \textcolor{black}{appeared} at different locations in the camera field of view suffer distinct distortions in the fisheye image. The location of the head in the fisheye image helps the estimation network determine the distortion information of the head region, and thus improves the accuracy of HPE. Another distinctive advantage of our method is that it does not require rectification, or the knowledge of the camera parameters of the fisheye camera. It significantly simplifies \textcolor{black}{HPE} computation. Considering there are no publicly accessible fisheye image datasets for head pose estimation, we created \textcolor{black}{a} fisheye-\textcolor{black}{distorted} version of the three popular public benchmark datasets, BIWI, 300W-LP, and AFLW2000, and performed comparison experiments on the new datasets. Experimental results show that our proposed approach achieved higher accuracy compared with two-stage and other one-stage methods. \textcolor{black}{We also constructed a real-world fisheye dataset to evaluate the effectiveness of the proposed method. Experimental results show that our method decreased the average estimation errors for real-world fisheye images compared with the two-stage methods.}

\section{Related Work}\label{relatedworks}

\textcolor{temp}{Over the last two decades, advances in hardware and software have made perspective (pinhole) cameras used in a wide variety of cognition concerned applications, including human-computer interaction and human-robot interaction. Many efficient methods based on perspective cameras were proposed to capture and estimate the head pose~\cite{wang2018human}, gesture~\cite{tcds202handpose}, pose of body~\cite{tcds2023humanpose,tcds2024humanpose} and gaze direction~\cite{tcds2023gaze} and achieved promising results. These research works significantly improved the development of intelligent robotic systems and artificial cognitive systems. }

\textcolor{black}{With the ability to capture image of a larger scene area than perspective (pinhole) camera, omnidirectional cameras have been widely employed in many applications including video surveillance~\cite{dinchangtseng_2017_1}, autonomous driving~\cite{yaozuye_2020_12}, and 3D reconstruction~\cite{kottari2019real}. Fisheye and catadioptric cameras are two typical omnidirectional camera types. Fisheye camera is a sort of conventional camera combined with a fisheye-shaped lens, while the catadioptric camera combines a standard camera with a shaped mirror. The drawback of catadioptric camera is the shaped mirror is fragile and the camera itself occludes the central part of the image, which makes catadioptric camera a less popular option than the fisheye camera.}

Fisheye \textcolor{black}{camera is} \textcolor{temp}{able to capture} images with an ultra-wide field-of-view (FOV). As high-resolution and portable fisheye cameras are becoming relatively affordable, more and more computer vision tasks, such as object detection~\cite{dinchangtseng_2017_1}, semantic segmentation~\cite{yaozuye_2020_12}, \textcolor{black}{fall detection~\cite{kottari2019real}, saliency prediction~\cite{zhu2021multiscale}}, face detection~\cite{cheng-yunyang_2021_15}, and face recognition~\cite{yi-chenglo_2022_16}, have gradually shifted from traditional rectilinear images to fisheye images. However, all existing \textcolor{black}{RGB image-based} HPE algorithms were designed for rectilinear images.

Current \textcolor{black}{RGB image-based} HPE methods for rectilinear images are commonly grouped into landmark-based methods and landmark-free methods. Landmark-based methods estimate head pose by investigating the geometric relations implied among facial landmarks, while landmark-free methods explore the whole face image and estimate the head pose directly from the input image.

Landmark-based methods detect the facial landmarks and estimate the 3D pose of an object from n 3D-to-2D point-correspondences with Perspective-n-Point (PnP) algorithms. \textcolor{black}{Both classical~\cite{burgos2013robust, cao2014face} and deep learning-based}~\cite{adrianbulat_2017_13, hongwenzhang_2018_23} and remarkably improved the accuracy of landmark-based head pose estimation. However, the accuracy of these methods is highly dependent on the correct prediction of the landmark positions. Hence, inferior landmark localization caused by occlusion and extreme rotation can consequently impair the accuracy of head pose estimation. 

Rather than exploring the geometric relations among facial landmarks, landmark-free methods estimate the head pose by learning the characteristics of the whole image. Benefitting from the advancement of machine learning techniques, especially the success of deep learning, landmark-free methods have achieved promising results on many widely used datasets and become a popular approach for head pose estimation. Hopenet employed a multi-loss ResNet50 network to predict Euler angles and achieved accurate results by training with both classification and regression losses~\cite{ruiz_2018}. FSA-Net proposed a feature aggregation method to improve the accuracy of head pose estimation~\cite{tsun-yiyang_2019_35}. Li et al. proposed a lightweight network to obtain highly accurate head pose estimation~\cite{xiaoli_2022_34}. They \textcolor{black}{preprocessed the image with} image rectification to reduce the negative effect of perspective distortion and employed a discriminative weighted loss to train the network. QuatNet proposed a Quaternion-based head pose regression framework and obtained a more effective HPE method~\cite{Hsu_2019}. Img2pose estimated the head pose by calculating six degrees of freedom (6DoF) and 3D face poses of the head directly from raw images without face detection or landmark localization~\cite{Albiero_2021}.
	
All \textcolor{black}{the aforementioned} HPE methods assume an ideal pinhole camera model is employed. \textcolor{black}{The pinhole camera model mimics the geometrical projection carried out by a pinhole camera and has a small FOV. However, wide FOV cameras violate the rules of perspective transformation and linear projection, and result in distortions in wide-angle image. Therefore, the performance of pinhole camera model degrades significantly for wide-angle image.  In order to characterize such distortion, Geyer et al. presented a unifying theory for all central catadioptric systems~\cite{geyer2000unifying}. A central catadioptric camera means the catadioptric camera equipped with hyperbolic, parabolic, or elliptical mirror. They proved that all central catadioptric systems can be modeled with a projection from a sphere to a plane where the projection center is on the sphere diameter and the plane is perpendicular to the sphere diameter. Furthermore, Ying et al. presented a unified imaging model for fisheye and central catadioptric cameras~\cite{ying2004can}. With this unified model, a fisheye image can be transformed into a central catadioptric one, and vice versa.}

\begin{figure}[tbp]
    \centering
    \includegraphics[width=3.2in]{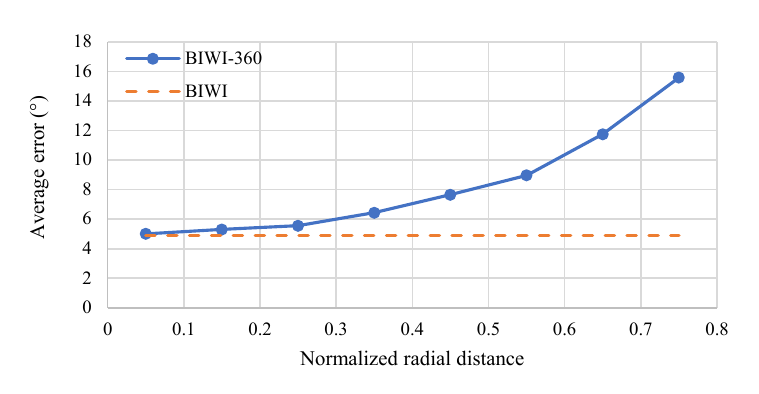}
    \caption{Average head pose estimation error in fisheye image versus normalized radial distance. The solid line is the result on the \textcolor{black}{fisheye-distorted} BIWI-360 dataset. The dotted line is the result on the \textcolor{black}{non-distorted} BIWI dataset.}
    \label{fig:figure2}
\end{figure}

Radial distortion is the main distortion in fisheye camera, which arises from the spherical shape of the lenses and causes the projection onto the image plane to be displaced from the ideal position along the radial axis~\cite{jinlongfan_2022_20}. As shown in Fig.~\ref{fig:figure1}, while the fisheye lens provides ultra-wide view, the face in the image is significantly distorted by the radial distortion of the lens. The radial distortion not only influences the appearance of the face, but also greatly degrades the performance of \textcolor{black}{the} existing HPE methods. Hopenet~\cite{ruiz_2018} is one of the best HPE models, which was trained on the 300W-LP dataset. We used their pre-trained model and tested it on the fisheye-\textcolor{black}{distorted} version of the BIWI-360 dataset we created. The testing results are shown in Fig.~\ref{fig:figure2}, where the horizontal coordinate means the normalized radial distance, which represents the distance of the face from the optical axis. The vertical coordinate \textcolor{black}{depicts} the \textcolor{temp}{average error of the estimated} Euler angles in degrees.
	
\textcolor{black}{Fig.~\ref{fig:figure2} shows} the head pose estimation performance of Hopenet~\cite{ruiz_2018} is quite good if the location of the head is closed to the center of the fisheye image, but it degrades as head moves toward the perimeter where the \textcolor{black}{distortion is worse}. The radial distortion introduced by the fisheye lens is closely related to the distance between the object and the optical axis. When the head is very close to the optical axis or the center of the fisheye image, the accuracy of the current HPE methods is quite good because the radial distortion is minimal. When the head is positioned away from the optical axis of the camera and the radial distortion is more evident, their performance suffers.

Literature review shows that learning related tasks at the same time can achieve better performance than training tasks \textcolor{black}{separately}~\cite{sebastianruder_2017_21}. Studies show that head pose estimation can also achieve better performance with the help of multi-task learning~\cite{rajeevranjan_2017_4,rajeevranjan_2019_62}. Ranjan presented an all-in-one convolutional neural network (CNN) for simultaneous face detection, face alignment, pose estimation, gender recognition, smile detection, age estimation, and face recognition~\cite{rajeevranjan_2017_4}. HyperFace first learns common features by CNN, and then simultaneously performs several tasks in a single framework: face detection, landmark localization, gender classification, and head pose estimation~\cite{rajeevranjan_2019_62}. \textcolor{black}{Both of these research work proposed} multi-task learning networks that combined the task of head pose estimation with other facial analysis tasks and obtained improved performance for the combined tasks.
	
To the best of our knowledge, there is \textcolor{black}{no} reported research \textcolor{black}{of} HPE for fisheye image. We believe that the accuracy of HPE for fisheye images can be improved if the location of the head in the image can be introduced into the estimation network. We develop an end-to-end CNN to estimate the head pose by jointly learning the head pose and head location. Experimental results show that our approach of using the knowledge of head location achieves higher accuracy compared to two-stage and other one-stage methods.

\section{Methods}\label{proposedmethod}
\subsection{Network Architecture}
\begin{figure*}[tbp]
    \centering
    \includegraphics[width=6.2in]{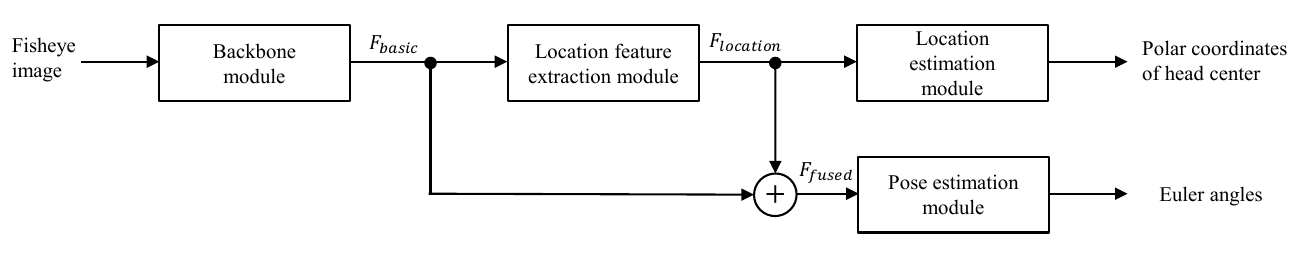}
    %\vspace{0.3cm}%插入垂直间隔
    \caption{The overview of the proposed network.}
    \label{fig:figure3}
\end{figure*}

\textcolor{black}{The main idea of this work is to use} the head location information to help the backbone module learn the camera distortion and improve the HPE accuracy. In this paper, we propose an end-to-end convolutional neural network to estimate the head pose from the fisheye image directly. The architecture of the proposed network, as shown in Fig.~\ref{fig:figure3}, consists of the backbone module, location feature extraction module, pose estimation module, and location estimation module.

In the proposed network, the backbone module is first used to extract features from the fisheye image. The obtained features are then fed to the location feature extraction module which learns the location-related features supervised by the ground truth of head location. We fuse the features learned from the backbone module and the location-related features to obtain the fused features. The fused features are fed to an estimation module to estimate the Euler angles of the head pose. 
	
\subsection{Backbone Module}
The backbone module is used to extract high-level features directly from the fisheye facial image. Considering the residual neural network (ResNet)~\cite{He_2016} is an excellent convolutional neural network that can solve the vanishing gradient problem in deep learning models, and is used in many promising HPE networks for feature learning~\cite{ruiz_2018,rajeevranjan_2019_62,inesrieger_2019_0,mingzhenshao_2019_2,binhuang_2020_1}, we adopt the ResNet50 structure as the backbone of our model. The employed ResNet50 is pre-trained on ImageNet-1k~\cite{Deng_2009} for the object recognition task. The input of the whole network is a single RGB image in size of 224×224. The size of the output feature maps ($\mathbf{F}_{\mathrm{basic}}$  in Fig.~\ref{fig:figure3}) from this module is 7×7×2048.

\subsection{Location Feature Extraction Module }
As the fisheye distortion of an object presented in the fisheye image is closely related to its location, estimating the location of the object can help the network learn the cues of distortion. We employ a location feature extraction module to learn the features of head location from $\mathbf{F}_{\mathrm{basic}}$. Specifically, the location feature extraction module \textcolor{black}{generates} the location-related features and provides high-level clues implying object distortion. We fuse the location-related features with the basic features and provide location guidance to the estimation of head pose. 

\begin{figure*}[tbp]
    \centering
    \includegraphics[width=6.2in]{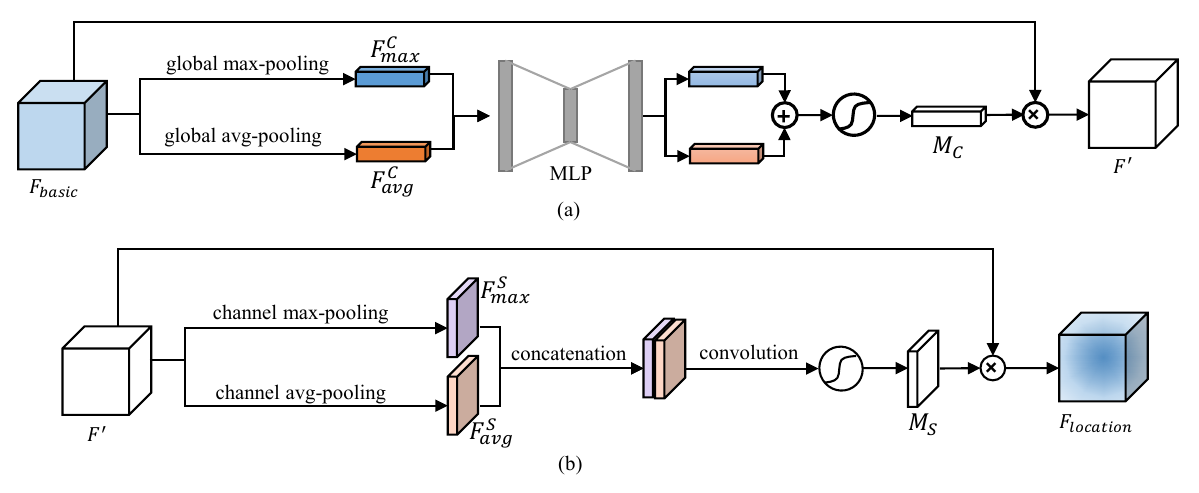}
    %\vspace{0.3cm}
    \caption{The location feature extraction module based on the attention mechanism in~\cite{sanghyunwoo_2018_5} that includes two sequential submodules: (a) submodule of channel attention and (b) submodule of spatial attention.}
    \label{fig:figure4}
\end{figure*}

To make the network focus on more important features for the task of head location estimation, as shown in Fig.~\ref{fig:figure4}, we apply an attention mechanism in the location feature extraction module, which includes a channel attention submodule and a spatial attention submodule successively~\cite{sanghyunwoo_2018_5}.
	
For the input feature $\mathbf{F}_{\mathrm{basic}}\in\mathbb{R}^{C \times H \times W }$, the channel attention submodule first generates two feature vectors by using both average-pooling and max-pooling operations, denoted as $\mathbf{F}_{\mathrm{avg}}^{\mathrm{C}}\in\mathbb{R}^{C}$ and $\mathbf{F}_{\mathrm{max}}^{\mathrm{C}}\in\mathbb{R}^{C}$.
\begin{equation}
    {\mathbf{F}_{\mathrm{avg}}^{\mathrm{C}}= \frac{1}{H \times W} \sum\limits_{h=0}^H \sum\limits_{w=0}^W \mathbf{F}_{\mathrm{basic}}(c,h,w)},
    \label{eq1}	
\end{equation}

\begin{equation}
    {\mathbf{F}_{\mathrm{max}}^{\mathrm{C}}=\max\limits_{h, w} \mathbf{F}_{\mathrm{basic}}(c, h, w)}
    \label{eq2}
\end{equation}

After the pooling operations, the two feature vectors are fed into a multi-layer perceptron (MLP) with one hidden layer to produce two weight vectors. The weight-matrixes of the hidden layer and the output layer are denoted as $\mathbf{W}_1\in\mathbb{R}^{(C/r)\times C}$ and  $\mathbf{W}_2\in\mathbb{R}^{C \times(C/r)}$, where $r$ is the reduction ratio and set to 16. The hidden layer is followed by a ReLU activation function for nonlinearity. Then the two weight vectors are added and activated by a sigmoid function to obtain the channel attention map $\mathbf{M}_{\mathrm{C}}\in\mathbb{R}^C$,	
\begin{equation}
        {\mathbf{M}_{\mathrm{C}}=\sigma\left(\mathbf{W}_{2}\delta\left(\mathbf{W}_{1}\mathbf{F}_{\mathrm{avg}}^{\mathrm{C}}\right)+\mathbf{W}_{2}\delta\left(\mathbf{W}_{1}\mathbf{F}_{\mathrm{max} }^{\mathrm{C}}\right)\right)
        }
    \label{eq3}
\end{equation}
where $\sigma(\cdot)$ and $\delta(\cdot)$ indicate the sigmoid and ReLU function, respectively. Finally, the channel-weighted feature, $\mathbf{F}^{\prime}\in\mathbb{R}^{C \times H \times W }$, is obtained by multiplying the input feature and the channel attention map.

The spatial attention submodule performs the average-pooling and max-pooling operations on the channel-weighted feature, and generates two feature maps denoted as $\mathbf{F}_{\mathrm{avg}}\in\mathbb{R}^{H \times W }$  and  $\mathbf{F}_{\mathrm{max}}\in\mathbb{R}^{H \times W }$.	
\begin{equation}
    {\mathbf{F}_{\mathrm{avg}}^{\mathrm{S}}(h, w)=\frac{1}{C} \sum_{c=1}^C \mathbf{F^{\prime}}(c, h, w)}
    \label{eq4}
\end{equation}

\begin{equation}
    {\mathbf{F}_{\mathrm{max}}^{\mathrm{S}}(h, w)=\max\limits_{c} \mathbf{F^{\prime}}(c, h, w)}
    \label{eq5}
\end{equation}

After the pooling operations, the two maps are concatenated along the channel axis, and then convolved by a 7×7 convolution layer, producing the spatial attention map $\mathbf{M}_{\mathrm{S}}\in\mathbb{R}^{H \times W }$,	
\begin{equation}
    {\mathbf{M}_{\mathrm{S}}=\sigma\left(f^{7 \times 7}\left(\left[\mathbf{F}_{\mathrm{avg}}^{\mathrm{S}} ; \mathbf{F}_{max}^{\mathrm{S}}\right]\right)\right)}
    \label{eq6}
\end{equation}	
where $f^{7 \times 7} (\cdot)$ a convolution operation with a 7×7 kernel. Finally, the channel-weighted feature, $\mathbf{F^{\prime}}$, is multiplied by the spatial attention map $\mathbf{M}_{\mathrm{S}}$  to produce $\mathbf{F}_{\mathrm{location}}$ .

\subsection{Location Estimation Module}
The location estimation module is used to generate the final estimation of head location in the image and help the module of location feature extraction to learn the cues of location-related distortion. We express the location of the head center with the polar angle and normalized radial distance in polar coordinates. We split the range of polar angle, $\theta \in$ [-180°, 180°] into 72 equal intervals, and the normalized radial distance, $\rho \in [0,0.99]$ into 66 equal intervals. This design turns head location estimation into a \textcolor{black}{task of classification}.

The location estimation module receives $\mathbf{F}_{\mathrm{location}}$ from the location feature extraction module as an input feature. A global average pooling layer is first applied to the feature of $\mathbf{F}_{\mathrm{location}}$ and produces a one-dimensional (1D) feature vector for two parallel branches to estimate the values of ${\theta}$ and ${\rho}$. Each branch includes a fully connected (FC) layer, a softmax layer, and an expectation operation. The softmax layer turns the output of the FC layer into probabilities of ${\theta}$ or ${\rho}$ in the corresponding intervals. The expectation operation generates the final estimation of ${\theta}$ and ${\rho}$ by \textcolor{black}{using} Eqs.~(\ref{eq7}) and~(\ref{eq8}), respectively,
\begin{equation}
    {\hat{\theta}=5 \sum_{i=1}^M p_i^\theta\left(i-\frac{M+1}{2}\right)},
    \label{eq7}
\end{equation}		

\begin{equation}
    {\hat{\rho}=0.015 \sum_{i=1}^N p_i^\rho\left(i-\frac{1}{2}\right)},
    \label{eq8}
\end{equation}		
where $p_i^{\theta}$  and $p_i^{\rho}$ are the probabilities of the estimated polar angle and normalized radial distance in the corresponding $i^{th}$ interval. In Eq.~(\ref{eq7}), 5 is the interval length of polar angle and M is the number of intervals for the polar angle. In Eq.~(\ref{eq8}), 0.015 is the interval length of normalized radial distance and N is the number of intervals for the normalized radial distance. The subtracted terms in both equations shift the interval index to the midpoint of the interval.

Although we estimate the values of polar angle and normalized radial distance in the training stage, we do not need to calculate them in the inference stage. We employ the location estimation module to help the location feature extraction module to learn the cues of location-related distortion in \textcolor{black}{the} fisheye image.  We add the feature of $\mathbf{F}_{\mathrm{location}}$ and $\mathbf{F}_{\mathrm{basic}}$ to form a fused feature $\mathbf{F}_{\mathrm{fused}}$ for the HPE and anticipate the feature of head location to help improve the performance of head pose estimation.

\subsection{Pose Estimation Module} 
The pose estimation module is used to generate the final estimation of head pose, which is expressed by Euler angles (pitch, yaw, roll). To facilitate the comparison with~\cite{ruiz_2018,tsun-yiyang_2019_35,xiaoli_2022_34}, we only considered samples whose Euler angles were within the range of [-99°, 99°]. The range of each Euler angle is split into 66 equal intervals of 3 degrees. This design turns head pose estimation into a \textcolor{black}{task of classification}.

\begin{figure}[ht]
    \centering
    \includegraphics[width=3.1in]{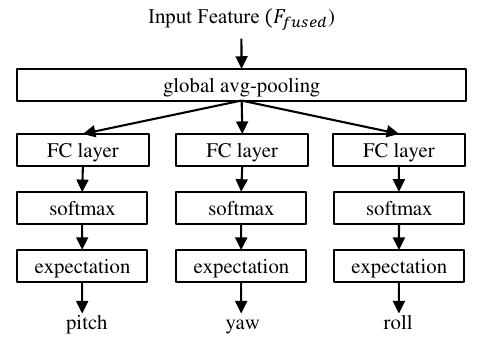}
    %\vspace{0.3cm}
    \caption{The structure of the pose estimation module.}
    \label{fig:figure5}
\end{figure}	

As shown in Fig.~\ref{fig:figure5}, a global average pooling layer is first applied to the feature of $\mathbf{F}_{\mathrm{fused}}$ and produces a one-dimensional (1D) feature vector for three parallel branches to estimate the values of pitch, yaw, and roll. Each branch employs an FC layer followed by a softmax function. The output of the softmax function is a 1D vector of 66 values that represent the probabilities that the Euler angle of head pose falls into the intervals. Then an expectation layer is used to generate the final estimation of each Euler angle by \textcolor{black}{using} Eqs.~(\ref{eq9}) to~(\ref{eq11}). 	
\begin{equation}
    {\widehat{pitch}= 3 \sum_{i=1}^N p_i^p\left(i-\frac{N+1}{2}\right)},
    \label{eq9}
\end{equation}	

\begin{equation}
    {\widehat{yaw}= 3 \sum_{i=1}^N p_i^y\left(i-\frac{N+1}{2}\right)},
    \label{eq10}
\end{equation}	

\begin{equation}
    {\widehat{roll}= 3 \sum_{i=1}^N p_i^r\left(i-\frac{N+1}{2}\right)},
    \label{eq11}
\end{equation}	 
where $p_i^p$,$p_i^y$,and  $p_i^r$ are the probabilities of the angles of pitch, yaw, and roll in the $i^{th}$ interval, respectively, \textcolor{black}{the value 3 before the symbol of summation} represents 3 degrees for each interval and N is the number of intervals for each Euler angle (N=66 in our design). The subtracted term shifts the interval index to the midpoint of the interval.

\subsection{Loss Funtion}
In our proposed multi-task learning network for pitch, yaw, roll, normalized radial distance and polar angle, the loss function for each task can be formulated \textcolor{black}{by} Eq.~(\ref{eq12}): 
\begin{equation}
    {\mathcal{L} = \text{CE}(y, \hat{y}) + \alpha \cdot \text{MSE}(y, \hat{y})},
    \label{eq12}
\end{equation}
where $\text{CE}(\cdot)$ and $\text{MSE}(\cdot)$ represent cross entropy and mean squared error loss functions, respectively. $y$ is the ground truth of the corresponding task, while $\hat{y}$ is the estimated result of the corresponding task. \textcolor{black}{We express the location of the head center with the polar angle and normalized radial distance in polar coordinates. The pitch, yaw, roll, and polar angle are estimated in degree, while the estimated value of radial distance is normalized into the range of [0, 0.99] and the range of polar angle is [-180°, 180°]. Experiments show the MSE of the estimated normalized radial distance differs from the MSEs of other estimated tasks by at least two orders of magnitude. Therefore, we set the value of $\alpha$ for the MSE loss of estimated normalized radial distance to 100, while the values of $\alpha$ for the MSE losses of estimated pitch, yaw, roll, and polar angle to 1.}

The overall loss is represented by a weighted sum of the loss for each task, as shown in Eq.~(\ref{eq13}):
\begin{equation}
\textcolor{black}{
    {\mathcal{L}_{total } = 
    \mathcal{L}_{pitch} +\mathcal{L}_{yaw }  + \mathcal{L}_{roll} + \lambda_1 \mathcal{L}_{\rho}  +\lambda_2 \mathcal{L}_{\theta}},}
\label{eq13}
\end{equation}
where $\mathcal{L}_{pitch }$, $\mathcal{L}_{yaw }$, $\mathcal{L}_{roll }$, $\mathcal{L}_{\rho}$ and $\mathcal{L}_{\theta}$ are the losses for pitch, yaw, roll, normalized radial distance and polar angle, respectively; \textcolor{black}{$\lambda_1$ and $\lambda_2$ are the weights to balance the losses for normalized radial distance and polar angle, respectively.}

\textcolor{black}{The training process of one iteration is summarized in Algorithm 1 for better understanding.}

\begin{algorithm}
    \setstretch{1.046}
    \caption{\textcolor{black}{Training Process in One Iteration}}
    \label{alg}
    \begin{algorithmic}[1] %这个1 表示每一行都显示数字
        \REQUIRE %算法的输入参数：Input
        \textcolor{black}{
          Fisheye image ($\boldsymbol{I}_{input}$), model parameters ($\delta$),\\
          \qquad learning rate ($\eta$), backbone module $(H_{BAC})$,\\
          \qquad location feature extraction module $(H_{LFE})$, \\
          \qquad location estimation module $(H_{LOE})$,\\
          \qquad pose estimation module $(H_{POE})$.
          }
        \ENSURE %算法的输出：Output
        \textcolor{black}{Updated model parameters $\delta^{*}$;}
        \STATE \textcolor{black}{Extract features from fisheye image,\\ $F_{basic} = H_{BAC}(\boldsymbol{I}_{input})$;}
        \STATE \textcolor{black}{Generate the location-related features,\\ $F_{location} = H_{LFE}(F_{basic})$;}
        \STATE \textcolor{black}{Generate the fused features,\\ $F_{fuesd}=F_{location}+F_{basic}$;}
        \STATE \textcolor{black}{Predict the polar coordinates,\\ $[\hat{\rho},\hat{\theta}]=H_{LOE}(F_{location})$;}
        \STATE \textcolor{black}{Predict the Euler angles,\\ $[\hat{pitch},\hat{yaw},\hat{roll}]=H_{POE}(F_{fuesd})$;}
        \STATE \textcolor{black}{Calculate the loss $\mathcal{L}_{total}$ via Eq. (13);}
        \STATE \textcolor{black}{Update model parameters,\\$\delta^{*}\gets \delta - \eta\frac{\partial }{\partial \delta}\mathcal{L}_{total} $.}
    \end{algorithmic}
\end{algorithm}

\section{Experiments and Discussion}\label{results}
As there is no publicly accessible benchmark dataset of fisheye images for head pose estimation, we created the fisheye-\textcolor{black}{distorted} version of three popular public benchmark datasets for our experiments. 
To demonstrate the superiority of our proposed method, we performed several comparative experiments with both two-stage and one-stage HPE approaches for fisheye image. Firstly, we compared the estimation accuracy of the proposed network with three two-stage approaches which rectify the fisheye image using a recently developed rectification method and perform head pose estimation using three state-of-the-art methods~\cite{ruiz_2018,tsun-yiyang_2019_35,xiaoli_2022_34} trained on the rectilinear datasets. These state-of-the-art methods were included for comparison because their goals are also to predict Euler angles from a single RGB image. Secondly, we compared the estimation accuracy of the proposed network with three existing state-of-the-art (one-stage) methods~\cite{ruiz_2018,tsun-yiyang_2019_35,xiaoli_2022_34}. For a fair comparison, these networks were retrained on the fisheye-\textcolor{black}{distorted} version of three popular head post estimation datasets we created. Thirdly, we evaluated the processing speed and number of parameters of the proposed network. The final experiment involved a series of ablation tests to investigate how head location guidance affects the performance of the proposed framework.

\subsection{Datasets}
We created datasets of fisheye images annotated with the ground truth of head pose to evaluate the performance of our proposed method. Using a fisheye camera to capture a large number of face images with their accompanying head pose ground truth is a daunting task. Instead, we selected three open-source popular head pose datasets, including BIWI~\cite{gabrielefanelli_2012_59}, 300W-LP~\cite{xiangyuzhu_2019_60}, and AFLW2000~\cite{xiangyuzhu_2019_60}, and created their fisheye-\textcolor{black}{distorted} version. We named these new datasets BIWI-360, 300W-LP-360, and AFLW2000-360, respectively. We followed the same mapping function used in~\cite{jianglinfu_2019_39,jianglinfu_2019_36,yi-hsinli_2021_4,tangweili_2020_38} to transform the original images in the datasets into fisheye-\textcolor{black}{distorted} images while keeping the ground truth of the head pose unchanged. 
 
\textbf{Source Datasets.} The BIWI dataset was collected in a lab environment. It contains 15,678 images collected by a Kinect v2 device~\cite{gabrielefanelli_2012_59}. Subjects (6 females and 14 males) moved and turned their heads to all possible angles observed from a frontal position. RGB images and depth information were included for each image and head pose annotations were created from the depth information. The head poses of these images are in the range of ±75° for yaw, ±60° for pitch, and ±50° for roll. 

Both the 300W-LP~\cite{xiangyuzhu_2019_60} and AFLW2000~\cite{xiangyuzhu_2019_60} datasets were created by cropping out the face region directly from photos or video frames downloaded from the Internet or other similar sources. Besides, both datasets use 3D Dense Face Alignment (3DDFA) to fit a morphable 3D model to 2D input images and provide accurate head poses as ground truth. 300W-LP additionally generates synthetic views, greatly expanding the number of images. More specifically, the 300W-LP dataset employed face profiling with a 3D image model to expand the 300W dataset which combines several face alignment datasets including LFPW~\cite{peternbelhumeur_2011_58}, AFW~\cite{xiangxinzhu_2012_57}, HELEN~\cite{erjinzhou_2013_56}, IBUG~\cite{christossagonas_2013_55}, and XMSVTS~\cite{kmesser_1999_54}, with 68 landmarks for each face image. As a result, 61,225 samples (16556 from LFPW~\cite{peternbelhumeur_2011_58}, 5,207 from AFW~\cite{xiangxinzhu_2012_57}, 37676 from HELEN~\cite{erjinzhou_2013_56}, and 1,786 from IBUG~\cite{christossagonas_2013_55}) were generated. 300WP-LP further flipped all these samples to double the sample size to 122,450. AFLW2000 is a very challenging dataset with large pose variations with various facial expressions and illumination conditions. AFLW2000 provides face images, corresponding head pose ground truth, and 3D face for the first 2000 images in AFLW~\cite{martinkostinger_2011_50}. 

\textbf{Mapping Function.} To visually approximate the \textcolor{black}{distortion of a} fisheye \textcolor{black}{lens}, we first normalized the coordinates of rectilinear image, which means that (0.0, 0.0) represents the origin of the coordinate system located at the image center, and (±1.0, ±1.0) represents the coordinates of the four quadrants. 
Let $(x,y)$ be the normalized coordinates of an arbitrary pixel in a rectilinear image. Its corresponding pixel $\left(x^{\prime}, y^{\prime}\right)$ in the fisheye image can be formulated as Eq.~(\ref{eq14}).
\begin{equation}
    {\left(x^{\prime}, y^{\prime}\right)=\left(x \sqrt{1-\frac{y^2}{2}}, 
    y \sqrt{1-\frac{x^2}{2}}\right)}.
    \label{eq14}	
\end{equation}
Furthermore, we introduce a coordinate scaling factor $e^{-\frac{r^2}{2}}$ to control the severity of its distortion \textcolor{black}{with} Eq.~(\ref{eq15}),
\begin{equation}
    {\left(x^{\prime \prime}, y^{\prime \prime}\right)=\left(x^{\prime} e^{-\frac{r^2}{2}}, y^{\prime} e^{-\frac{r^2}{2}}\right)},
    \label{eq15}		
\end{equation}	
where $r$ is the radial distance of the pixel $\left(x^{\prime}, y^{\prime}\right)$ to the image center~\cite{jianglinfu_2019_36}.

\textcolor{black}{This mapping function introduces fisheye distortion, which manifests itself as an effect of “squeezing” the content towards the perimeter of the fisheye image. Different from the distortion in real-world fisheye images, which involves various camera parameters or lens distortion parameters, the mapping function we adopted is not camera or lens specific.}

\begin{figure*}[!t]
    \centering
    \subfigure[]{
        \includegraphics[width=2.1in]{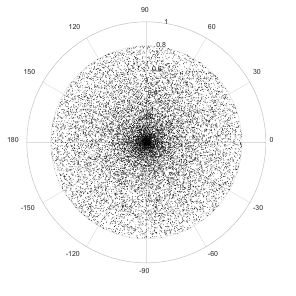}        
      }
    \subfigure[]{
        \includegraphics[width=2.1in]{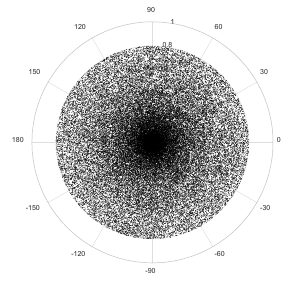}
        }	
    \subfigure[]{
        \includegraphics[width=2.1in]{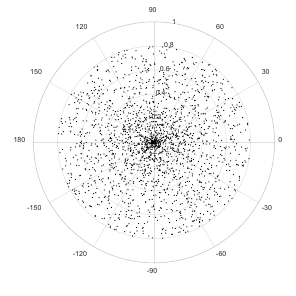}
        }
    \caption{The distribution of the face locations in polar coordinates for different fisheye datasets. (a) BIWI-360, (b) 300WP-LP-360, and (c) AFLW2000-360.}
    \label{fig:figure6}
\end{figure*}

\begin{figure*}[tbp]%[!t]
    \centering
    \subfigure[]{
        \includegraphics[width=2.1in]{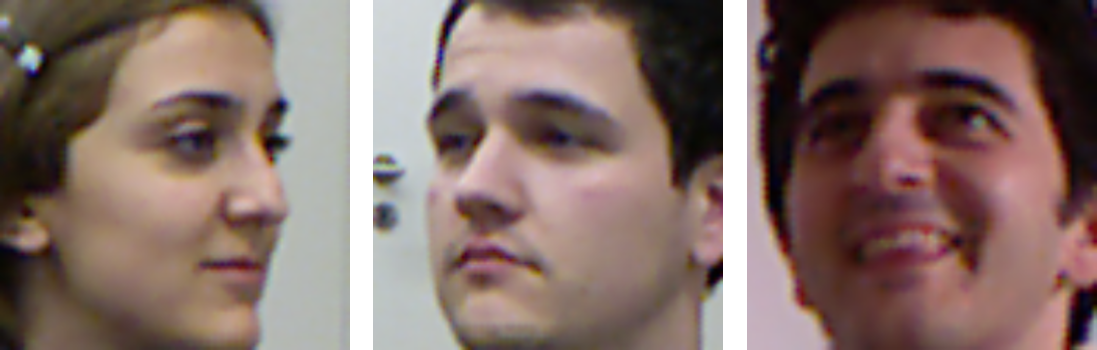}        
      }
    \subfigure[]{
        \includegraphics[width=2.1in]{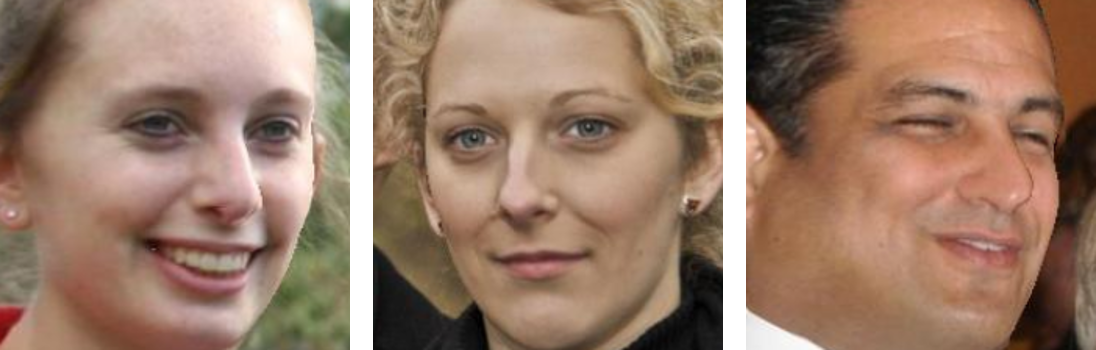}
        }	
    \subfigure[]{
        \includegraphics[width=2.1in]{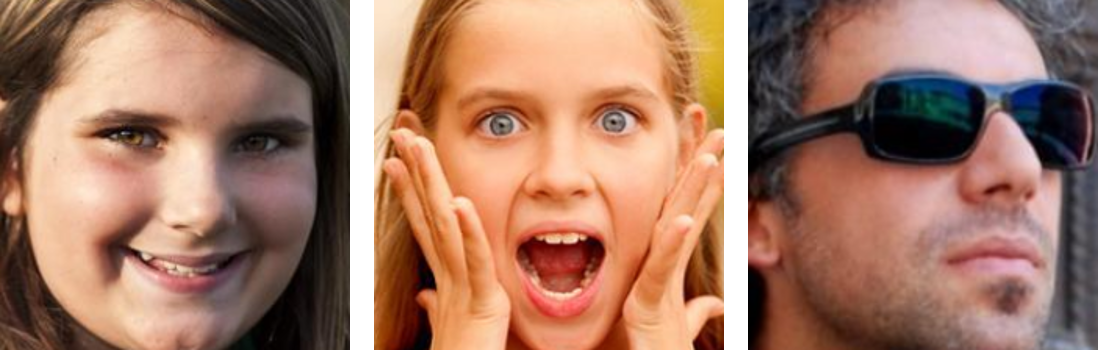}
        }
    \subfigure[]{
        \includegraphics[width=2.1in]{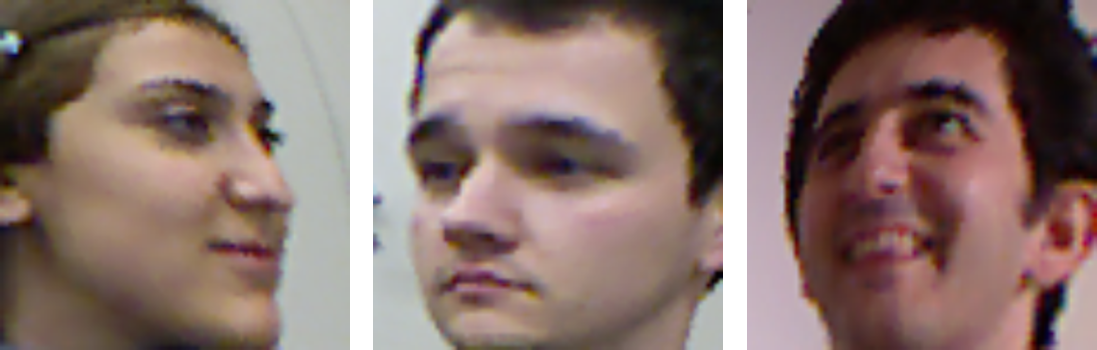}        
      }
    \subfigure[]{
        \includegraphics[width=2.1in]{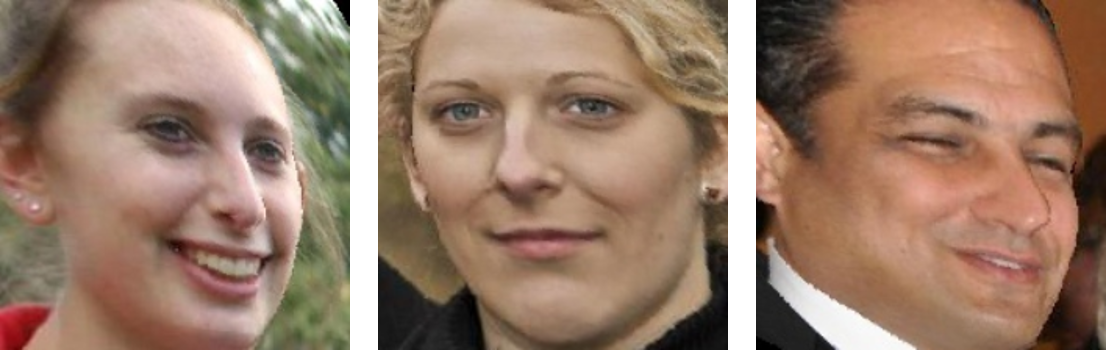}
        }	
    \subfigure[]{
        \includegraphics[width=2.1in]{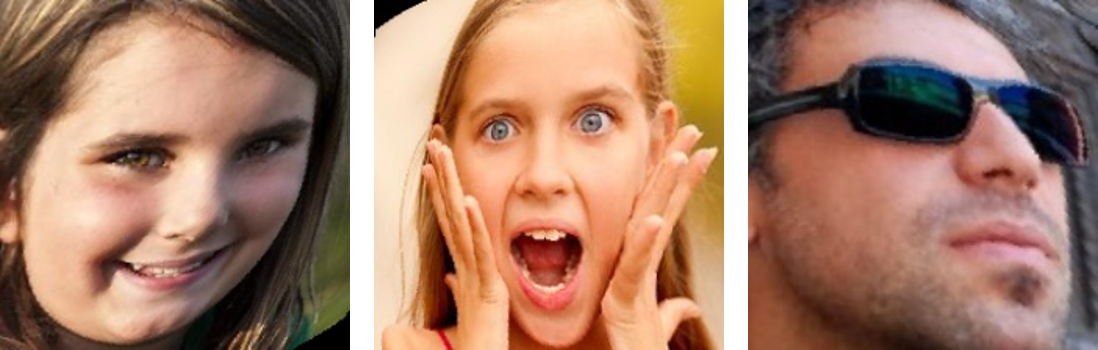}
        }
    \caption{\textcolor{black}{Examples of face images from different datasets. The upper row represents original (rectilinear) face images from different datasets (a) BIWI, (b) 300W-LP, (c) AFLW2000, and the bottom row represents the corresponding fisheye face images from the synthesized datasets (d) BIWI-360, (e) 300W-LP-360, (f) AFLW2000-360.}}
    \label{fig:figure7}
\end{figure*}	

\textbf{Synthetized Datasets.} In the source datasets, the regions of faces usually appear around the center of image. However, in real applications, the region of face may appear elsewhere in a photo or video frame. The face region subjects neglectable geometric distortion if it is very close to the center of fisheye image. The degree of distortions increases as the face region moves towards the perimeter. To reflect the geometric distortion induced at different locations, a good fisheye dataset must have sufficient samples of which the face region may appear at all possible locations.

For each rectilinear image in the source datasets, we represented the location of face by the center of its bounding box. Then we created a canvas and replaced a canvas region with a facial region from the rectilinear image. In our work, the width and height of the canvas were both set to 5 times of the face bounding box. Next, we set up a polar coordinate system as the fisheye distortion is usually circularly symmetrical \textcolor{black}{with respect to} the image center~\cite{Zhao_2022}. The polar point was set at the canvas center and the angular angle increases counterclockwise. The radial distance of face was normalized by \textcolor{black}{half} the width of canvas and was sampled from the uniform distribution (0, 0.8). The polar angle of face was sampled from the uniform distribution (-180, 180). Finally, we transformed the canvas using the mapping function in Eqs.~(\ref{eq14}) and~(\ref{eq15}) to create the effect of fisheye distortion. The corresponding bounding box and landmarks were also converted into the fisheye image coordinate system. Once the coordinate transformation was completed, we cropped the face areas from the resulting fisheye image. 

We performed the above operations on the BIWI, 300W-LP, and AFLW2000 datasets, and correspondingly obtained three synthesized fisheye datasets with varying face locations in the images. The distributions of the face locations in polar coordinates for these synthetic fisheye datasets are shown in Fig.~\ref{fig:figure6}. Each dot in these maps represents the face location in polar coordinates of each sample in the corresponding datasets. \textcolor{black}{In order to have an intuitive impression on the synthesized fisheye datasets, we show some samples of face image from the original and the corresponding synthetic datasets in Fig.~\ref{fig:figure7}.}

\subsection{Configuration and Evaluation Metrics}
In all our experiments, the proposed method was implemented on the Pytorch platform. Cuda 10.2 was also employed to speed up model training. Our network was trained for 25 epochs using Adam optimization~\cite{diederikpkingma_2014_53} with an initial learning rate of $10^{-4}$, $\beta_1 = 0.9$, and $\beta_2 = 0.9$. The learning rate was reduced at the $10^{th}$ epoch and the $20^{th}$ epoch by a factor of 0.5 each. Each color channel was normalized using the mean and standard deviation of ImageNet before training and testing. Face detection was performed using RetinaFace~\cite{jiankangdeng_2019_51}.

Two evaluation metrics were used in all experiments. One was the average error in the yaw, pitch, and roll angles in degrees. The other metric was the mean absolute error (MAE), which is expressed in Eq.~(\ref{eq16}).
\begin{equation}
    {\text{MAE}=\frac{1}{3N} \sum_{k=1}^{N}\left(\left|\alpha_k-\hat{\alpha}_k\right|+\left|\beta_k-\hat{\beta}_k\right|+\left|\gamma_k-\hat{\gamma}_k\right|\right)},
    \label{eq16}
\end{equation}
where N is the number of samples in the testing dataset, $(\alpha_k, \beta_k, \gamma_k)$ are the ground truth of the Euler angles and $(\hat{\alpha}_k, \hat{\beta}_k, \hat{\gamma}_k)$ are the estimated Euler angles.

\subsection{Comparison with two-stage approaches}
In this subsection, we compared the performance of our proposed method with the two-stage approaches. The two-stage approaches for comparison employed the progressively complementary network (PCN)~\cite{Yang_2021} to rectify the fisheye image first, and then estimated the head pose from the rectified image with the state-of-the-art HPE methods for rectilinear images~\cite{ruiz_2018,tsun-yiyang_2019_35,xiaoli_2022_34}. PCN was adopted as it is independent of the camera parameters and can generate an undistorted image from the fisheye image directly~\cite{Yang_2021}. We used the well-trained models of the two-stage approaches as reported in their corresponding publications. The estimation results obtained by using the two-stage approaches are reported as “original methods + PCN” in \textcolor{black}{our experiments}.

Our network was trained on the 300W-LP-360 dataset while others were trained on the 300W-LP dataset. For training our network, the batch size was set to 128, and the weights \textcolor{black}{$\lambda_1$ and $\lambda_2$ in Eq.~(\ref{eq13}) were set to 5 and 0.001}, respectively. In testing, our network used the fisheye image as input, while others used the rectified fisheye image as input. Testing was performed on two datasets: AFLW2000-360 and BIWI-360 datasets. The experimental results of testing on the AFLW2000-360 dataset are shown in Table~\ref{tab:4}. The experimental results of testing on the BIWI-360 dataset are shown in Table~\ref{tab:5}.

\begin{table}[ht]\small
    \renewcommand{\arraystretch}{1.2}
    \setlength{\tabcolsep}{1.4mm}
    \centering 
    \caption{Average error comparisons on the AFLW2000-360 dataset. Our network was trained on the 300W-LP-360 dataset while others were trained on the 300W-LP dataset.}
    \begin{tabular}{ccccc}
        %\hline\noalign{\smallskip}
        \toprule 
        \textbf{Method} & \textbf{Yaw} & \textbf{Pitch} & \textbf{Roll} & \textbf{MAE} \\ 	
        \midrule
        Hopenet ($\alpha=1$)~\cite{ruiz_2018} + PCN  & 6.31° & 8.20° & 6.20° & 6.90° \\
        Hopenet ($\alpha=2$)~\cite{ruiz_2018} + PCN  & 6.68° & 8.36° & 6.10° & 7.05° \\
        FSA-Net~\cite{tsun-yiyang_2019_35} + PCN     & 4.97° & 7.66° & 5.04° & 5.89° \\
        Li et al.~\cite{xiaoli_2022_34} + PCN        & 5.39° & 7.68° & 5.23° & 6.10° \\
        Ours & \textbf{3.65°} & \textbf{5.85°} & \textbf{4.17°} & \textbf{4.55°} \\
        \bottomrule[1pt]
    \end{tabular}
    \label{tab:4}
\end{table}
% \vspace{-5 mm}
\begin{table}[ht]\small
    \renewcommand{\arraystretch}{1.2}
    \setlength{\tabcolsep}{1.4mm}
    \centering 
    \caption{Average error comparisons on the BIWI-360 dataset. Our network was trained on the 300W-LP-360 dataset while others were trained on the 300W-LP dataset.}
    \begin{tabular}{ccccc}
        %\hline\noalign{\smallskip}
        \toprule 
        \textbf{Method} & \textbf{Yaw} & \textbf{Pitch} & \textbf{Roll} & \textbf{MAE} \\ 	
        \midrule
        Hopenet ($\alpha=1$)~\cite{ruiz_2018} + PCN  & 6.21° & 9.84° & 4.26° & 6.77° \\
        Hopenet ($\alpha=2$)~\cite{ruiz_2018} + PCN  & 6.98° & 7.82° & 4.32° & 6.37° \\
        FSA-Net~\cite{tsun-yiyang_2019_35} + PCN     & 6.37° & 8.34° & 4.28° & 6.33° \\
        Li et al.~\cite{xiaoli_2022_34} + PCN        & 6.77° & 8.13° & 4.50° & 6.47° \\
        Ours & \textbf{4.53°} & \textbf{7.24°} & \textbf{3.73°} & \textbf{5.16°} \\
        \bottomrule[1pt]
    \end{tabular}
    \label{tab:5}
\end{table}

As shown in Tables~\ref{tab:4} and~\ref{tab:5}, our proposed network obtained much lower average errors in yaw, pitch, roll angles, and MAE than the two-stage methods. Experiments demonstrated that the proposed method achieved better performance compared to \textcolor{black}{other} two-stage approaches.

\subsection{Comparison with one-stage approaches}
Head pose estimators designed for rectilinear image can be retrained to deal with fisheye distortion as the one-stage approaches. We retrained three existing state-of-the-art methods for comparison: Hopenet~\cite{ruiz_2018}, FSA-Net~\cite{tsun-yiyang_2019_35}, and the method proposed by Li et al.~\cite{xiaoli_2022_34}, and cited them as “original methods + retrain” in experiment. We used the same configurations of the compared methods, including model architectures, loss function, and hyper-parameters, as reported in their corresponding publications. The experimental results are listed in Tables~\ref{tab:1} to~\ref{tab:3}. 

Tables~\ref{tab:1} to~\ref{tab:3} show the average errors in degrees with different experimental protocols. We followed the experimental protocol proposed in~\cite{sankhasmukherjee_2015_52} and randomly split the BIWI-360 dataset into 70\% (16 videos) for training and 30\% (8 videos) for testing to obtain the results shown in Table~\ref{tab:1}. We repeated this process three times and reported the average measurement error. In this protocol, the batch size was set to 32, and the weights \textcolor{black}{$\lambda_1$ and $\lambda_2$ in Eq.~(\ref{eq13}) were set to 10 and 0.001}, respectively. We also performed training on the 300W-LP-360 dataset while testing on the AFLW2000-360 dataset and the BIWI-360 dataset, respectively. We followed the same settings as Hopenet~\cite{ruiz_2018} and only considered samples whose Euler angles were within the range of [-99°, 99°]. In this protocol, the batch size was set to 128, and the weights \textcolor{black}{$\lambda_1$ and $\lambda_2$ in Eq.~(\ref{eq13}) were set to 5 and 0.001}, respectively. Results of these experiments are shown in Tables~\ref{tab:2} and~\ref{tab:3}. 

\begin{table}[ht]\small
    \renewcommand{\arraystretch}{1.2}
    \setlength{\tabcolsep}{1.4mm}
    \centering
    \caption{Comparison of average errors using the BIWI-360 dataset for training and testing.}
    \begin{tabular}{ccccc}
        %\hline\noalign{\smallskip}
        \toprule 
        \textbf{Method} & \textbf{Yaw} & \textbf{Pitch} & \textbf{Roll} & \textbf{MAE} \\ 	
        \midrule
        Hopenet ($\alpha=1$)~\cite{ruiz_2018} + retrain  & 2.99° & 3.50° & 3.26° & 3.25°  \\
        FSA-Net~\cite{tsun-yiyang_2019_35} + retrain     & 4.08° & 5.14° & 4.26° & 4.50° \\
        Li et al.~\cite{xiaoli_2022_34} + retrain        & 3.39° & 4.60° & 3.35° & 3.78° \\
        Ours & \textbf{2.66°} & \textbf{3.11°} & \textbf{2.87°} & \textbf{2.88°} \\
        \bottomrule[1pt]
    \end{tabular}
    \label{tab:1} 
\end{table}
% \vspace{-5 mm}
\begin{table}[ht]\small
    \renewcommand{\arraystretch}{1.2}
    \setlength{\tabcolsep}{1.4mm}
    \centering
    \caption{Comparison of average errors using the 300W-LP-360 dataset for training and the AFLW2000-360 dataset for testing.}
    \begin{tabular}{ccccc}
        %\hline\noalign{\smallskip}
        \toprule 
        \textbf{Method} & \textbf{Yaw} & \textbf{Pitch} & \textbf{Roll} & \textbf{MAE} \\ 	
        \midrule
        Hopenet ($\alpha=1$)~\cite{ruiz_2018} + retrain  & 3.60° & 6.05°  & 4.42° & 4.69° \\
        Hopenet ($\alpha=2$)~\cite{ruiz_2018} + retrain  & \textbf{3.59°} & 5.96° & 4.49° & 4.68° \\
        FSA-Net~\cite{tsun-yiyang_2019_35} + retrain     & 5.61° & 7.16°  & 5.70° & 6.16° \\
        Li et al.~\cite{xiaoli_2022_34} + retrain        & 4.19° & 6.29°  & 4.49° & 4.99° \\
        Ours & 3.65° & \textbf{5.85°} & \textbf{4.17°} & \textbf{4.55°} \\
        \bottomrule[1pt]
    \end{tabular}
    \label{tab:2} 
\end{table}
\vspace{-5 mm}
\begin{table}[h]\small
    \renewcommand{\arraystretch}{1.2}
    \setlength{\tabcolsep}{1.4mm}
    \centering 
    \caption{Comparison of average errors using the 300W-LP-360 dataset for training and the BIWI-360 dataset for testing.}
    \begin{tabular}{ccccc}
        %\hline\noalign{\smallskip}
        \toprule 
        \textbf{Method} & \textbf{Yaw} & \textbf{Pitch} & \textbf{Roll} & \textbf{MAE} \\ 	
        \midrule
        Hopenet ($\alpha=1$)~\cite{ruiz_2018} + retrain  & 5.04° & \textbf{7.22}° & 3.77° & 5.35° \\
        Hopenet ($\alpha=2$)~\cite{ruiz_2018} + retrain  & 5.22° & 7.51° & 3.77° & 5.50° \\
        FSA-Net~\cite{tsun-yiyang_2019_35} + retrain     & 6.18° & 9.53° & 4.58° & 6.76° \\
        Li et al.~\cite{xiaoli_2022_34} + retrain        & 5.07° & 7.99° & 4.10° & 5.72° \\
        Ours & \textbf{4.53°} & 7.24° & \textbf{3.73°} & \textbf{5.16°}\\
        \bottomrule[1pt]
    \end{tabular}
    \label{tab:3}
\end{table}
    
Table~\ref{tab:1} shows our method obtained the best performance among all compared methods when used the BIWI-360 dataset for training and testing. The training for Tables~\ref{tab:2} and~\ref{tab:3} were performed on the 300W-LP-360 dataset. As shown in Table~\ref{tab:2}, although our method had a slightly higher average error in yaw angle than the retrained Hopenet in the testing on the AFLW2000-360 dataset, our method obtained the best estimation for pitch, roll angles, and MAE. Moreover, when tested on the BIWI-360 dataset (Table~\ref{tab:3}), our method \textcolor{black}{performed better} in estimating yaw, roll angles, and MAE, despite having a comparable error in pitch angle compared to the retrained Hopenet.
 
The above results show that our approach achieved better or comparable performance compared to other one-stage approaches. As we used head location estimation as the auxiliary task for head pose estimation, these experiments demonstrate the knowledge of head location improves the performance of our network for head pose estimation.

\subsection{Processing Speed}
We also conducted an experiment to evaluate the processing speed of our network. We compared our network with two-stage methods \textcolor{black}{and one-stage methods} on a GPU platform. The GPU platform employed a single NVIDIA GeForce TITAN X. The version of CUDA library was 10.2. For a fair comparison, all methods were tested on the same target platform. Each method was evaluated 100 times. The measured average runtime is shown in \textcolor{black}{Tables~\ref{tab:6} and~\ref{tab:7}}. 
 
\textcolor{black}{As shown in Table~\ref{tab:6}}, our method runs faster and uses a smaller number of parameters compared with the two-stage methods because our network does not require a separate image rectification process. \textcolor{black}{\ref{tab:7} shows that our network runs slower than other one-stage approaches because our network includes a location branch and additional computational cost. Nevertheless, the processing speed of our method is good enough for real-world applications in most cases.}

\begin{table}[!t]\small
    \renewcommand{\arraystretch}{1.2}
    \setlength{\tabcolsep}{3.2mm}
    \centering 
    \caption{\textcolor{black}{The comparison of processing speed and number of parameters for our network and two-stage methods.} }
    \begin{tabular}{ccc}
        %\hline\noalign{\smallskip}
        \toprule 
        \textbf{Method} & \textbf{Speed (fps)} & \textbf{\#Params}  \\ 	
        \midrule
        Hopenet~\cite{ruiz_2018} + PCN             & 18 & 59.56M \\
        FSA-Net~\cite{tsun-yiyang_2019_35} + PCN   & 19 & 36.81M \\
        Li et al.~\cite{xiaoli_2022_34} + PCN      & 21 & 35.85M \\
        Ours & \textbf{61} & \textbf{28.92M}  \\
        \bottomrule[1pt]
    \end{tabular}
    \label{tab:6}
\end{table}

\begin{table}[!t]\small
    % \arrayrulecolor{blue}%增加
    \renewcommand{\arraystretch}{1.2}
    \setlength{\tabcolsep}{3.2mm}
    \centering 
    \caption{\textcolor{black}{The comparison of processing speed and number of parameters for our network and one-stage methods.} }
    \begin{tabular}{ccc}
        %\hline\noalign{\smallskip}
        \toprule 
        \textcolor{black}{\textbf{Method}} & \textcolor{black}{\textbf{Speed (fps)}} & \textcolor{black}{\textbf{\#Params} } \\ 	
        \midrule
        \textcolor{black}{Hopenet~\cite{ruiz_2018} + retrain}              & \textcolor{black}{66}  & \textcolor{black}{23.92M} \\
        \textcolor{black}{FSA-Net~\cite{tsun-yiyang_2019_35} + retrain}    & \textcolor{black}{84}  & \textcolor{black}{1.17M} \\
        \textcolor{black}{Li et al.~\cite{xiaoli_2022_34} + retrain}       & \textcolor{black}{123} & \textcolor{black}{0.21M} \\
        \textcolor{black}{Ours} & \textcolor{black}{61} & \textcolor{black}{28.92M}  \\
        \bottomrule[1pt]
    \end{tabular}
    \label{tab:7}
\end{table}

\subsection{Ablation Study}
Experiments were conducted to show the effectiveness of our proposed head location guided head pose estimation on fisheye distortion. All experiments used the BIWI-360 dataset for training (70\%) and testing (30\%). We repeated this process three times and reported the average measurement error. The experimental results are shown in Table~\ref{tab:8}. In Table~\ref{tab:8}, $\mathcal{L}_{\rho}$ denotes the loss function for normalized radial distance used in training. $\mathcal{L}_{\theta}$ represents the loss function for polar angle used in training. 

\textcolor{black}{To further verify the impact of location guidance, we performed an ablation study in every possible configuration, i.e., the network is equipped with or without the location feature extraction module, the location supervisions for polar angle $\mathcal{L}_{\theta}$, and the location supervisions for normalized radial distance $\mathcal{L}_{\rho}$. The experimental results are listed in Table ~\ref{tab:8}.}

\begin{table}[ht]\small
    % \arrayrulecolor{color}%增加
    \renewcommand{\arraystretch}{1.2} %1.2倍行高
    \setlength{\tabcolsep}{1mm} %设置每个单元格的宽度
    \centering 
    \caption{\textcolor{black}{The effectiveness of the location guidance of head for HPE. The BIWI-360 dataset was used for training (70\%) and testing (30\%). “\checkmark” and “\mbox{-}” indicate with or without the corresponding module/supervision.}}
    \begin{tabular}{ccccccc}
        %\hline\noalign{\smallskip}
        \toprule 	
        \multirow{2}{*}{Location feature} & \multicolumn{2}{c}{Location supervision} &  
        \multirow{2}{*}{Yaw} &  \multirow{2}{*}{Pitch} &  \multirow{2}{*}{Roll} &  \multirow{2}{*}{MAE}\\
        \cmidrule{2-3}       extraction module  & \quad $\mathcal{L}_{\rho}$ &  \quad $\mathcal{L}_{\theta}$ \\
        \midrule
        \mbox{-}  &  \mbox{-}  &   \mbox{-} & 2.72° & 3.41°  & 3.06°  & 3.06° \\
        \textcolor{black}{\mbox{-}}  & \textcolor{black}{\mbox{-}}   & \textcolor{black}{\checkmark} & \textcolor{black}{2.71°} & \textcolor{black}{3.25°}  & \textcolor{black}{2.98°}  & \textcolor{black}{2.98°} \\
        \textcolor{black}{\mbox{-}}  & \textcolor{black}{\checkmark} & \textcolor{black}{\mbox{-}}   & \textcolor{black}{2.59°} & \textcolor{black}{3.36°}  & \textcolor{black}{2.94°}  & \textcolor{black}{2.96°} \\
        \textcolor{black}{\mbox{-}}  & \textcolor{black}{\checkmark} & \textcolor{black}{\checkmark} & \textcolor{black}{2.66°} & \textcolor{black}{3.31°}  & \textcolor{black}{3.03°}  & \textcolor{black}{3.00°} \\            
        \checkmark &  \mbox{-}  &   \mbox{-} & 2.74° & 3.28°  & 3.14°  & 3.06° \\
        \checkmark &  \mbox{-}  & \checkmark & \textbf{2.55°} & 3.27°  & 2.97°  & 2.93° \\
        \checkmark & \checkmark &  \mbox{-}  & 2.56° & 3.25°  & 2.95°  & 2.92° \\
        \checkmark & \checkmark & \checkmark & 2.66° & \textbf{3.11°}  & \textbf{2.87°}  & \textbf{2.88°}\\        
        \bottomrule[1pt]
    \end{tabular}
    \label{tab:8} 
\end{table}	

\textcolor{black}{The first four rows of Table~\ref{tab:8} show that using the guidance of head location, although without the location feature extraction module, helped improve accuracy of the network. The first and fifth rows of Table~\ref{tab:8} show that the location feature extraction module without location supervision would not be beneficial in improving the estimation accuracy of the network. The last three rows of Table~\ref{tab:8} show that the combination of head location supervision and location feature extraction module contributed to more accurate estimation results. The first and last rows} of Table~\ref{tab:8} show that the network using both location feature extraction module and head location had an across-the-board improvement and 5.88\% lower MAE compared with the network without the knowledge of head location. 
 
\subsection{\textcolor{black}{Studies on Hyperparameters}}
\begin{figure}[hbp]
    \centering
    \includegraphics[width=3.2in]{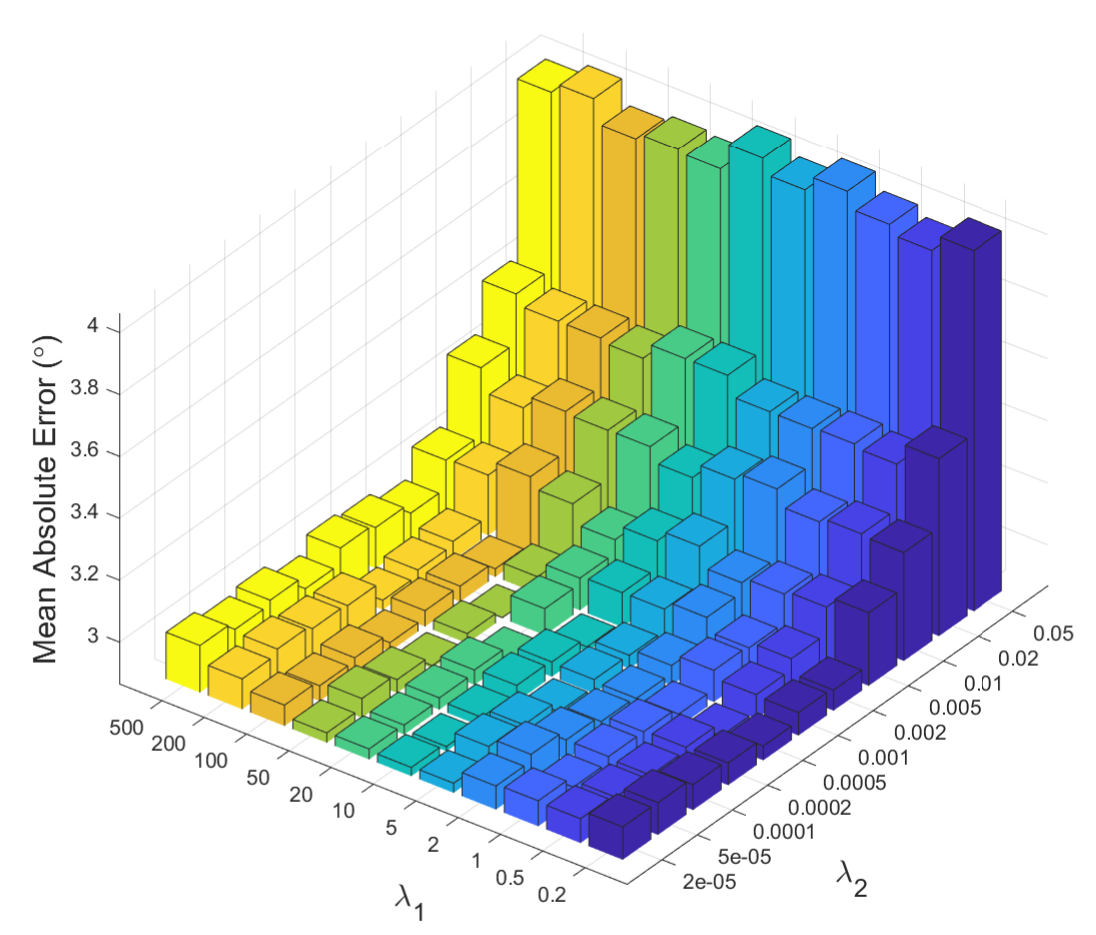}
    \caption{\textcolor{black}{The mean absolute error of Euler angles (in degrees) with different $\lambda_1$ and $\lambda_2$ on the BIWI-360 dataset. We use logarithmic scales to show the values of $\lambda_1$ and $\lambda_2$.}}
    \label{fig:figure8}
\end{figure}

\textcolor{black}{Experiments were conducted to explore the sensitivity of our proposed approach for the hyperparameters $\lambda_1$ and $\lambda_2$. We tested the performance of our proposed network by setting each of $\lambda_1$ and $\lambda_2$ to 11 different values. Specifically, we set $\lambda_1$ to 0.2, 0.5, 1, 2, 5, 10, 20, 50, 100, 200, 500, and $\lambda_2$ to 0.00002, 0.00005, 0.0001, 0.0002, 0.0005, 0.001, 0.002, 0.005, 0.01, 0.02, 0.05 in the experiment. All experiments used the BIWI-360 dataset for training (70\%) and testing (30\%). We repeated this process three times and reported the average measurement error. The experimental results are visualized in Fig.~\ref{fig:figure8}.}

\textcolor{black}{The experimental results show that the best performance was achieved by setting $\lambda_1$ between 10 and 20, and $\lambda_2$ between 0.0005 and 0.001. When $\lambda_1$ and $\lambda_2$ were set to 10 and 0.001, the obtained mean absolute error of Euler angles was 2.88°.}

\subsection{\textcolor{black}{Real-World Head Pose Estimation}}
\begin{figure}[tbp]
    \centering
    \includegraphics[width=3.2in]{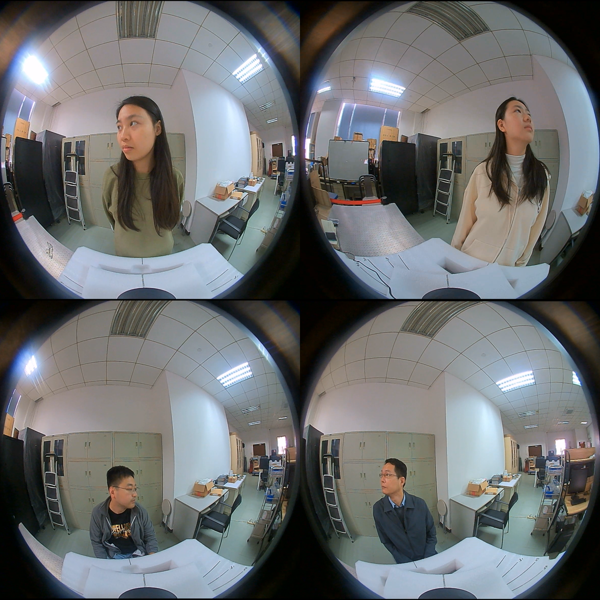}
    \caption{\textcolor{black}{Some samples of real-world fisheye images captured by JR$^\circledR$HF900.}}
    \label{fig:figure9}
\end{figure}	

\textcolor{black}{To evaluate the performance of our proposed method on real fisheye images, we constructed a real-world dataset with fisheye images captured by a calibrated fisheye camera (JR$^\circledR$HF900). The focal length and FOV of this camera are 1.1mm and 210°, respectively. The resolution of the captured fisheye image is 1080$\times$1080p. The dataset contains 70 sequences of 10 different people (7 men and 3 women, 5 wearing glasses) recorded while sitting or standing about 0.5 meters away from the fisheye camera. Some sample frames of sequences captured by the fisheye camera are shown in Fig.~\ref{fig:figure9}. The subjects rotated their heads to various positions. We captured videos of each subject from seven different angles, so that the subjects could appear in all possible locations within the fisheye images. More specifically, we kept the camera's optical axis to be approximately horizontal, and then used a camera gimbal to turn the fisheye camera on the horizontal plane so that the angle of the optical axis was at 0°, ±15°, ±30°, and ±45° to the direction of subject’s face. For each fisheye image, we employed RetinaFace~\cite{jiankangdeng_2019_51} to detect the face region and discarded the fisheye images when face detection failed to detect the face.}

\textcolor{black}{To annotate the head pose in the real-world fisheye image, we used the Camera Calibrator App~\cite{scaramuzza2006toolbox} provided by MATLAB R2021b to estimate the parameters of our fisheye camera, and then used these parameters to decrease the image distortion induced from lens of the fisheye camera. Finally, an excellent rectilinear-image-based HPE method [24] was employed to estimate the head pose from the rectified image. The Euler angles estimated by the HPE method was regarded as the ground truth of the head pose in the real-world fisheye image. As a result, the real-world dataset contains 21,815 fisheye images, annotated with head pose in Euler angles. The head pose range covers about ±90° for yaw, ±60° for pitch, and ±45° for roll.}

\begin{figure*}[!t]
    \centering
    \includegraphics[width=6.4in]{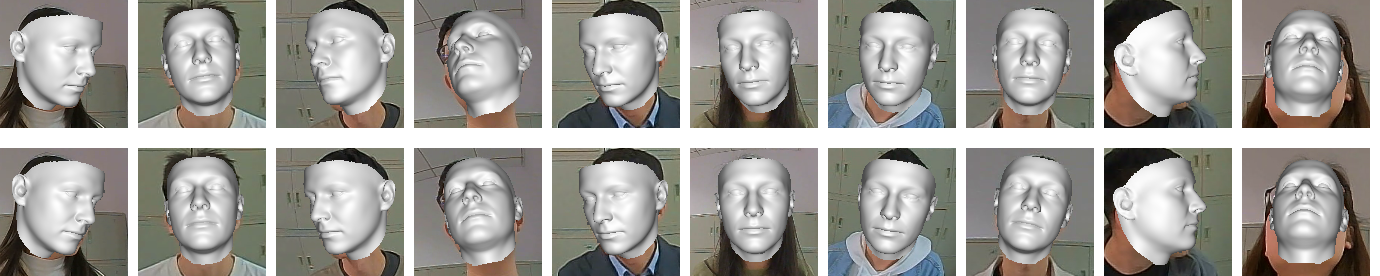}
    %\vspace{0.3cm}%插入垂直间隔
    \caption{\textcolor{black}{Comparison of the visualized ground truth of real-world samples (the upper row) and the head pose estimated by our proposed method (the bottom row). Head poses are visualized by the tools from~\cite{guo2020towards}.}}
    \label{fig:figure10}
\end{figure*}	

\begin{table}[ht]\small
% \arrayrulecolor{blue}%增加
\renewcommand{\arraystretch}{1.2}
\setlength{\tabcolsep}{1.4mm}
\centering 
\caption{\textcolor{black}{Comparison of our proposed method and two-stage methods on real-world fisheye images in the case no camera parameters are provided.}}
\begin{tabular}{ccccc}
    %\hline\noalign{\smallskip}
    \toprule 
    \textcolor{black}{\textbf{Method}} & \textcolor{black}{\textbf{Yaw}} & \textcolor{black}{\textbf{Pitch}} & \textcolor{black}{\textbf{Roll}} & \textcolor{black}{\textbf{MAE}} \\ 	
    \midrule
    \textcolor{black}{Hopenet ($\alpha=1$)~\cite{ruiz_2018} + PCN}  & \textcolor{black}{31.04°} & \textcolor{black}{17.57°} & \textcolor{black}{9.52°} & \textcolor{black}{19.37°} \\
    \textcolor{black}{Hopenet ($\alpha=2$)~\cite{ruiz_2018} + PCN}  & \textcolor{black}{28.97°} & \textcolor{black}{17.71°} & \textcolor{black}{10.74°} & \textcolor{black}{19.14°} \\
    \textcolor{black}{FSA-Net~\cite{tsun-yiyang_2019_35} + PCN}     & \textcolor{black}{15.85°} & \textcolor{black}{18.66°} & \textcolor{black}{8.11°} & \textcolor{black}{14.21°} \\
    \textcolor{black}{Li et al.~\cite{xiaoli_2022_34} + PCN}        & \textcolor{black}{26.64°} & \textcolor{black}{17.49°} & \textcolor{black}{8.33°} & \textcolor{black}{17.49°} \\
    \textcolor{black}{Ours} & \textcolor{black}{\textbf{9.23°}} & \textcolor{black}{\textbf{8.12°}} & \textcolor{black}{\textbf{6.37°}}& \textcolor{black}{\textbf{7.91°}} \\
    \bottomrule[1pt]
\end{tabular}
\label{tab:9}
\end{table}

\textcolor{black}{We evaluated our proposed method on the constructed real-world dataset and provides qualitatively demonstrated the performance of our method in Fig.~\ref{fig:figure10}. The upper row of Fig.~\ref{fig:figure10} shows the visualized ground truth of some real-world samples and the bottom row shows the head poses estimated by our method. These samples were captured from different angles and include all subjects. This visualized comparison shows that the poses estimated by our method are close to the ground truth, indicating the effectiveness of our approach with fisheye images of faces appearing at different locations.}

\textcolor{black}{We also compared the performances of our proposed method with two-stage methods on real-world fisheye images in the case that no camera parameters are provided. These experimental results are listed in Table~\ref{tab:9}. Experimental results show that our proposed method achieved better performance compared to the two-stage approaches in the case that no camera parameters are provided.}

\textcolor{black}{It is worth emphasizing that all experiments we performed, including on the synthetic and real-world datasets, assume a horizontal optical axis for the camera, which means the fisheye images in these datasets are mostly collected from a frontal view. In another common application scenario, ceiling-mounted fisheye cameras are used to capture top-view fisheye images~\cite{dinchangtseng_2017_1,kottari2019real}.}

\textcolor{black}{Although our proposed algorithm does not require the input image to be a front-view fisheye image, it will meet great challenges when the input is a top-view fisheye image for two reasons. First, the proposed method can hardly detect the presence of human face from the top view of a person. We employ a process of face detection for frontal view images to help the algorithm focus on the region of human face. For a fisheye image taken by a ceiling-mounted camera, the performance of the employed face detection algorithm will be affected dramatically. It may make the proposed method miss the head region appeared in the fisheye image. Second, the proposed method has little knowledge of the mapping between head pose and the top-view image of human.}

\textcolor{black}{As we use the synthetized fisheye datasets for the network training, while the synthetized fisheye datasets come from the datasets acquired by the camera with horizontal or nearly horizontal optical axis, the trained network only learned the knowledge about the head pose based on the fisheye image from the frontal view. We believe the knowledge of the proposed method comes from the datasets used in training. Datasets composed of top-view fisheye images together with the corresponding labels of head pose would help our proposed algorithm learn the knowledge of head pose from fisheye images taken by ceiling-mounted cameras.}
    
\section{Conclusion}\label{conclusion}
Head pose estimation from the fisheye image remains a challenging issue because of serious fisheye distortion in the peripheral region of the image. We propose an approach using the knowledge of \textcolor{temp}{head} location to reduce the impact of radial distortion for head pose estimation. We present an end-to-end convolutional neural network to estimate the head pose by proposing multi-task learning of head location and head pose. Another distinctive advantage of our method is that it does not require the process of rectification or calibration. Four experiments were designed to evaluate the performance of our approach. Experimental results show that the knowledge of the head location in the image \textcolor{black}{contributes} to \textcolor{black}{the} improved HPE accuracy. Our network outperforms the two-stage methods on two synthetic fisheye datasets and the one-stage methods on three synthetic fisheye datasets. Additionally, our network has a faster speed compared with the two-stage methods that involve a separate image rectification process.

\section*{Acknowledgments}
This work was supported by National Natural Science Foundation of China (62173353, 62171207), Science and Technology Program of Guangzhou, China (202007030011, 2023B03J1327), \textcolor{black}{and Guangdong Science and Technology Program (2023A1111120012)}.

\makeatletter
\def\changeBibColor#1{%	
\in@{#1}{
    %nawalalioua_2016_40,
    %boon-giinlee_2012_41,
    %murphy2010head,
    benfold2009guiding,
    george2020gazeroomlock,
    wang2018human,
    odobez2007cognitive,
    seemann2004head,
    tcds202handpose
    tcds2023humanpose,
    tcds2024humanpose,
    tcds2023gaze,
}% list of colored bib items
\ifin@\color{temp}\else\normalcolor\fi}
\xpatchcmd\@bibitem
{\item}
{\changeBibColor{#1}\item}
{}{\fail}
\xpatchcmd\@lbibitem
{\item}
{\changeBibColor{#2}\item}
{}{\fail}
\makeatother

% \makeatletter
% \def\changeBibColor#1{%	
% \in@{#1}{hansen2007scale,
%         kottari2019real,
%         zhu2021multiscale,
%         geyer2000unifying,
%         ying2004can,
%         cruz2012scale,
%         demonceaux2011central,
%         delibasis2018efficient,
%         georgakopoulos2018pose,
%         guo2020towards,
%         scaramuzza2006toolbox，
%         burgos2013robust,
%         cao2014face,
% }% list of colored bib items
% \ifin@\color{blue}\else\normalcolor\fi}
% \xpatchcmd\@bibitem
% {\item}
% {\changeBibColor{#1}\item}
% {}{\fail}
% \xpatchcmd\@lbibitem
% {\item}
% {\changeBibColor{#2}\item}
% {}{\fail}
% \makeatother

\bibliography{reference}

% Generated by IEEEtran.bst, version: 1.14 (2015/08/26)
\begin{thebibliography}{10}
\providecommand{\url}[1]{#1}
\csname url@samestyle\endcsname
\providecommand{\newblock}{\relax}
\providecommand{\bibinfo}[2]{#2}
\providecommand{\BIBentrySTDinterwordspacing}{\spaceskip=0pt\relax}
\providecommand{\BIBentryALTinterwordstretchfactor}{4}
\providecommand{\BIBentryALTinterwordspacing}{\spaceskip=\fontdimen2\font plus
\BIBentryALTinterwordstretchfactor\fontdimen3\font minus
  \fontdimen4\font\relax}
\providecommand{\BIBforeignlanguage}[2]{{%
\expandafter\ifx\csname l@#1\endcsname\relax
\typeout{** WARNING: IEEEtran.bst: No hyphenation pattern has been}%
\typeout{** loaded for the language `#1'. Using the pattern for}%
\typeout{** the default language instead.}%
\else
\language=\csname l@#1\endcsname
\fi
#2}}
\providecommand{\BIBdecl}{\relax}
\BIBdecl

\bibitem{khalilkhan_2021_0}
K.~Khan, R.~U. Khan, R.~Leonardi, P.~Migliorati, and S.~Benini, ``Head pose
  estimation: A survey of the last ten years,'' \emph{Signal Processing: Image
  Communication}, vol.~99, p. 116479, 2021.

\bibitem{chih-weichen_2011_6}
C.-W. Chen and H.~Aghajan, ``Multiview social behavior analysis in work
  environments,'' in \emph{2011 Fifth ACM/IEEE International Conference on
  Distributed Smart Cameras}.\hskip 1em plus 0.5em minus 0.4em\relax IEEE,
  2011, pp. 1--6.

\bibitem{nawalalioua_2016_40}
N.~Alioua, A.~Amine, A.~Rogozan, A.~Bensrhair, and M.~Rziza, ``Driver head pose
  estimation using efficient descriptor fusion,'' \emph{EURASIP Journal on
  Image and Video Processing}, vol. 2016, no.~1, pp. 1--14, 2016.

\bibitem{benfold2009guiding}
B.~Benfold and I.~Reid, ``Guiding visual surveillance by tracking human
  attention.'' in \emph{BMVC}, vol.~2, no.~6, 2009, p.~7.

\bibitem{george2020gazeroomlock}
C.~George, D.~Buschek, A.~Ngao, and M.~Khamis, ``Gazeroomlock: Using gaze and
  head-pose to improve the usability and observation resistance of 3d passwords
  in virtual reality,'' in \emph{Augmented Reality, Virtual Reality, and
  Computer Graphics: 7th International Conference, AVR 2020, Lecce, Italy,
  September 7--10, 2020, Proceedings, Part I 7}.\hskip 1em plus 0.5em minus
  0.4em\relax Springer, 2020, pp. 61--81.

\bibitem{wang2018human}
K.~Wang, R.~Zhao, and Q.~Ji, ``Human computer interaction with head pose, eye
  gaze and body gestures,'' in \emph{2018 13th IEEE International Conference on
  Automatic Face \& Gesture Recognition (FG 2018)}.\hskip 1em plus 0.5em minus
  0.4em\relax IEEE, 2018, pp. 789--789.

\bibitem{odobez2007cognitive}
J.-M. Odobez and S.~Ba, ``A cognitive and unsupervised map adaptation approach
  to the recognition of the focus of attention from head pose,'' in \emph{2007
  IEEE International Conference on Multimedia and Expo}, 2007, pp. 1379--1382.

\bibitem{hansen2007scale}
P.~Hansen, P.~Corke, W.~Boles, and K.~Daniilidis, ``Scale-invariant features on
  the sphere,'' in \emph{2007 IEEE 11th International Conference on Computer
  Vision}.\hskip 1em plus 0.5em minus 0.4em\relax IEEE, 2007, pp. 1--8.

\bibitem{cruz2012scale}
J.~Cruz-Mota, I.~Bogdanova, B.~Paquier, M.~Bierlaire, and J.-P. Thiran, ``Scale
  invariant feature transform on the sphere: Theory and applications,''
  \emph{International journal of computer vision}, vol.~98, pp. 217--241, 2012.

\bibitem{demonceaux2011central}
C.~Demonceaux, P.~Vasseur, and Y.~Fougerolle, ``Central catadioptric image
  processing with geodesic metric,'' \emph{Image and Vision Computing},
  vol.~29, no.~12, pp. 840--849, 2011.

\bibitem{delibasis2018efficient}
K.~K. Delibasis, ``Efficient implementation of gaussian and laplacian kernels
  for feature extraction from ip fisheye cameras,'' \emph{Journal of Imaging},
  vol.~4, no.~6, p.~73, 2018.

\bibitem{georgakopoulos2018pose}
S.~V. Georgakopoulos, K.~Kottari, K.~Delibasis, V.~P. Plagianakos, and
  I.~Maglogiannis, ``Pose recognition using convolutional neural networks on
  omni-directional images,'' \emph{Neurocomputing}, vol. 280, pp. 23--31, 2018.

\bibitem{tcds202handpose}
W.~Pang, Q.~Gao, Y.~Zhao, Z.~Ju, and J.~Hu, ``Basicnet: Lightweight 3d hand
  pose estimation network based on biomechanical structure information for
  dexterous manipulator teleoperation,'' \emph{IEEE Transactions on Cognitive
  and Developmental Systems}, 2022.

\bibitem{tcds2023humanpose}
G.~Wang, H.~Zeng, Z.~Wang, Z.~Liu, and H.~Wang, ``Motion projection consistency
  based 3d human pose estimation with virtual bones from monocular videos,''
  \emph{IEEE Transactions on Cognitive and Developmental Systems}, vol.~15, pp.
  784--793, 2023.

\bibitem{tcds2024humanpose}
Q.~Wu, Y.~Zhang, L.~Zhang, and H.~Yu, ``Parallel self-attention and
  spatial-attention fusion for human pose estimation and running movement
  recognition,'' \emph{IEEE Transactions on Cognitive and Developmental
  Systems}, vol.~16, no.~1, pp. 358--368, 2024.

\bibitem{tcds2023gaze}
Z.~Zhu, D.~Zhang, C.~Chi, M.~Li, and D.-J. Lee, ``A complementary dual-branch
  network for appearance-based gaze estimation from low-resolution facial
  image,'' \emph{IEEE Transactions on Cognitive and Developmental Systems},
  vol.~15, no.~3, pp. 1323--1334, 2023.

\bibitem{dinchangtseng_2017_1}
D.~Tseng, T.~I. of~Computer~Science, C.~U. Information~Engineering, C.~Chen,
  and C.~Tseng, ``Automatic detection and tracking in multi-fisheye cameras
  surveillance,'' \emph{International Journal of Computer and Electrical
  Engineering}, vol.~9, no.~1, pp. 370--383, 2017.

\bibitem{yaozuye_2020_12}
Y.~Ye, K.~Yang, K.~Xiang, J.~Wang, and K.~Wang, ``Universal semantic
  segmentation for fisheye urban driving images,'' in \emph{2020 IEEE
  International Conference on Systems, Man, and Cybernetics (SMC)}.\hskip 1em
  plus 0.5em minus 0.4em\relax IEEE, 2020, pp. 648--655.

\bibitem{kottari2019real}
K.~N. Kottari, K.~K. Delibasis, and I.~G. Maglogiannis, ``Real-time fall
  detection using uncalibrated fisheye cameras,'' \emph{IEEE transactions on
  cognitive and developmental systems}, vol.~12, no.~3, pp. 588--600, 2019.

\bibitem{zhu2021multiscale}
D.~Zhu, Y.~Chen, D.~Zhao, Y.~Zhu, Q.~Zhou, G.~Zhai, and X.~Yang, ``Multiscale
  brain-like neural network for saliency prediction on omnidirectional
  images,'' \emph{IEEE Transactions on Cognitive and Developmental Systems},
  vol.~14, no.~2, pp. 507--518, 2021.

\bibitem{cheng-yunyang_2021_15}
C.-Y. Yang and H.~H. Chen, ``Efficient face detection in the fisheye image
  domain,'' \emph{IEEE Transactions on Image Processing}, vol.~30, pp.
  5641--5651, 2021.

\bibitem{yi-chenglo_2022_16}
Y.-C. Lo, C.-C. Huang, Y.-F. Tsai, I.-C. Lo, A.-Y.~A. Wu, and H.~H. Chen,
  ``Face recognition for fisheye images,'' in \emph{2022 IEEE International
  Conference on Image Processing (ICIP)}.\hskip 1em plus 0.5em minus
  0.4em\relax IEEE, 2022, pp. 146--150.

\bibitem{burgos2013robust}
X.~P. Burgos-Artizzu, P.~Perona, and P.~Doll{\'a}r, ``Robust face landmark
  estimation under occlusion,'' in \emph{Proceedings of the IEEE international
  conference on computer vision}, 2013, pp. 1513--1520.

\bibitem{cao2014face}
X.~Cao, Y.~Wei, F.~Wen, and J.~Sun, ``Face alignment by explicit shape
  regression,'' \emph{International journal of computer vision}, vol. 107, pp.
  177--190, 2014.

\bibitem{adrianbulat_2017_13}
A.~Bulat and G.~Tzimiropoulos, ``How far are we from solving the 2d \& 3d face
  alignment problem?(and a dataset of 230,000 3d facial landmarks),'' in
  \emph{Proceedings of the IEEE international conference on computer vision},
  2017, pp. 1021--1030.

\bibitem{hongwenzhang_2018_23}
H.~Zhang, Q.~Li, and Z.~Sun, ``Joint voxel and coordinate regression for
  accurate 3d facial landmark localization,'' in \emph{2018 24th International
  Conference on Pattern Recognition (ICPR)}.\hskip 1em plus 0.5em minus
  0.4em\relax IEEE, 2018, pp. 2202--2208.

\bibitem{ruiz_2018}
N.~Ruiz, E.~Chong, and J.~M. Rehg, ``Fine-grained head pose estimation without
  keypoints,'' in \emph{Proceedings of the IEEE Conference on Computer Vision
  and Pattern Recognition Workshops (CVPRW)}, 2018, pp. 2074--2083.

\bibitem{tsun-yiyang_2019_35}
T.-Y. Yang, Y.-T. Chen, Y.-Y. Lin, and Y.-Y. Chuang, ``Fsa-net: Learning
  fine-grained structure aggregation for head pose estimation from a single
  image,'' in \emph{Proceedings of the IEEE/CVF Conference on Computer Vision
  and Pattern Recognition (CVPR)}, 2019, pp. 1087--1096.

\bibitem{xiaoli_2022_34}
X.~Li, D.~Zhang, M.~Li, and D.-J. Lee, ``Accurate head pose estimation using
  image rectification and a lightweight convolutional neural network,''
  \emph{IEEE Transactions on Multimedia}, 2022.

\bibitem{Hsu_2019}
H.-W. Hsu, T.-Y. Wu, S.~Wan, W.~H. Wong, and C.-Y. Lee,
  ``\BIBforeignlanguage{en}{Quatnet: Quaternion-based head pose estimation with
  multiregression loss},'' \emph{\BIBforeignlanguage{en}{IEEE Transactions on
  Multimedia}}, vol.~21, no.~4, p. 1035–1046, 2019.

\bibitem{Albiero_2021}
V.~Albiero, X.~Chen, X.~Yin, G.~Pang, and T.~Hassner, ``img2pose: Face
  alignment and detection via 6dof, face pose estimation,'' in
  \emph{Proceedings of the IEEE/CVF Conference on Computer Vision and Pattern
  Recognition (CVPR)}, 2021, pp. 7617--7627.

\bibitem{geyer2000unifying}
C.~Geyer and K.~Daniilidis, ``A unifying theory for central panoramic systems
  and practical implications,'' in \emph{Computer Vision-ECCV 2000: 6th
  European Conference on Computer Vision Dublin, Ireland, June 26--July 1, 2000
  Proceedings, Part II 6}.\hskip 1em plus 0.5em minus 0.4em\relax Springer,
  2000, pp. 445--461.

\bibitem{ying2004can}
X.~Ying and Z.~Hu, ``Can we consider central catadioptric cameras and fisheye
  cameras within a unified imaging model,'' in \emph{Computer Vision-ECCV 2004:
  8th European Conference on Computer Vision, Prague, Czech Republic, May
  11-14, 2004. Proceedings, Part I 8}.\hskip 1em plus 0.5em minus 0.4em\relax
  Springer, 2004, pp. 442--455.

\bibitem{jinlongfan_2022_20}
J.~Fan, J.~Zhang, S.~J. Maybank, and D.~Tao, ``Wide-angle image rectification:
  A survey,'' \emph{International Journal of Computer Vision}, vol. 130, no.~3,
  pp. 747--776, 2022.

\bibitem{sebastianruder_2017_21}
S.~Ruder, ``An overview of multi-task learning in deep neural networks,''
  \emph{arXiv preprint arXiv:1706.05098}, 2017.

\bibitem{rajeevranjan_2017_4}
R.~Ranjan, S.~Sankaranarayanan, C.~D. Castillo, and R.~Chellappa, ``An
  all-in-one convolutional neural network for face analysis,'' in \emph{2017
  12th IEEE international conference on automatic face \& gesture recognition
  (FG 2017)}.\hskip 1em plus 0.5em minus 0.4em\relax IEEE, 2017, pp. 17--24.

\bibitem{rajeevranjan_2019_62}
R.~Ranjan, V.~M. Patel, and R.~Chellappa, ``Hyperface: A deep multi-task
  learning framework for face detection, landmark localization, pose
  estimation, and gender recognition,'' \emph{IEEE Transactions on Pattern
  Analysis and Machine Intelligence}, vol.~41, no.~1, pp. 121--135, 2019.

\bibitem{He_2016}
K.~He, X.~Zhang, S.~Ren, and J.~Sun, ``Deep residual learning for image
  recognition,'' in \emph{Proceedings of the IEEE Conference on Computer Vision
  and Pattern Recognition Workshops (CVPRW)}, 2016, pp. 770--778.

\bibitem{inesrieger_2019_0}
I.~Rieger, T.~Hauenstein, S.~Hettenkofer, and J.-U. Garbas, ``Towards real-time
  head pose estimation: Exploring parameter-reduced residual networks on
  in-the-wild datasets,'' in \emph{Advances and Trends in Artificial
  Intelligence. From Theory to Practice: 32nd International Conference on
  Industrial, Engineering and Other Applications of Applied Intelligent
  Systems, IEA/AIE 2019, Graz, Austria, July 9--11, 2019, Proceedings
  32}.\hskip 1em plus 0.5em minus 0.4em\relax Springer, 2019, pp. 123--134.

\bibitem{mingzhenshao_2019_2}
M.~Shao, Z.~Sun, M.~Ozay, and T.~Okatani, ``Improving head pose estimation with
  a combined loss and bounding box margin adjustment,'' in \emph{2019 14th IEEE
  International Conference on Automatic Face \& Gesture Recognition (FG
  2019)}.\hskip 1em plus 0.5em minus 0.4em\relax IEEE, 2019, pp. 1--5.

\bibitem{binhuang_2020_1}
B.~Huang, R.~Chen, W.~Xu, and Q.~Zhou, ``Improving head pose estimation using
  two-stage ensembles with top-k regression,'' \emph{Image and Vision
  Computing}, vol.~93, p. 103827, 2020.

\bibitem{Deng_2009}
J.~Deng, W.~Dong, R.~Socher, L.-J. Li, K.~Li, and L.~Fei-Fei, ``Imagenet: A
  large-scale hierarchical image database,'' in \emph{2009 IEEE conference on
  computer vision and pattern recognition}.\hskip 1em plus 0.5em minus
  0.4em\relax Ieee, 2009, pp. 248--255.

\bibitem{sanghyunwoo_2018_5}
S.~Woo, J.~Park, J.-Y. Lee, and I.~S. Kweon, ``Cbam: Convolutional block
  attention module,'' in \emph{Proceedings of the European conference on
  computer vision (ECCV)}, 2018, pp. 3--19.

\bibitem{gabrielefanelli_2012_59}
G.~Fanelli, M.~Dantone, J.~Gall, A.~Fossati, and L.~Van~Gool, ``Random forests
  for real time 3d face analysis,'' \emph{International Journal of Computer
  Vision}, vol. 101, no.~3, pp. 437--458, 2012.

\bibitem{xiangyuzhu_2019_60}
X.~Zhu, X.~Liu, Z.~Lei, and S.~Z. Li, ``Face alignment in full pose range: A 3d
  total solution,'' \emph{IEEE Transactions on Pattern Analysis and Machine
  Intelligence}, vol.~41, no.~1, pp. 78--92, 2019.

\bibitem{jianglinfu_2019_39}
J.~Fu, S.~R. Alvar, I.~Bajic, and R.~Vaughan, ``Fddb-360: Face detection in
  360-degree fisheye images,'' in \emph{2019 IEEE Conference on Multimedia
  Information Processing and Retrieval (MIPR)}.\hskip 1em plus 0.5em minus
  0.4em\relax IEEE, 2019, pp. 15--19.

\bibitem{jianglinfu_2019_36}
J.~Fu, I.~V. Baji{\'c}, and R.~G. Vaughan, ``Datasets for face and object
  detection in fisheye images,'' \emph{Data in brief}, vol.~27, p. 104752,
  2019.

\bibitem{yi-hsinli_2021_4}
Y.-H. Li, I.-C. Lo, and H.~H. Chen, ``Deep face rectification for 360°
  dual-fisheye cameras,'' \emph{IEEE Transactions on Image Processing},
  vol.~30, pp. 264--276, 2021.

\bibitem{tangweili_2020_38}
T.~Li, G.~Tong, H.~Tang, B.~Li, and B.~Chen, ``Fisheyedet: A self-study and
  contour-based object detector in fisheye images,'' \emph{IEEE Access},
  vol.~8, pp. 71\,739--71\,751, 2020.

\bibitem{peternbelhumeur_2011_58}
P.~N. Belhumeur, D.~W. Jacobs, D.~J. Kriegman, and N.~Kumar, ``Localizing parts
  of faces using a consensus of exemplars,'' \emph{IEEE transactions on pattern
  analysis and machine intelligence}, vol.~35, no.~12, pp. 2930--2940, 2013.

\bibitem{xiangxinzhu_2012_57}
X.~Zhu and D.~Ramanan, ``Face detection, pose estimation, and landmark
  localization in the wild,'' in \emph{2012 IEEE conference on computer vision
  and pattern recognition}.\hskip 1em plus 0.5em minus 0.4em\relax IEEE, 2012,
  pp. 2879--2886.

\bibitem{erjinzhou_2013_56}
E.~Zhou, H.~Fan, Z.~Cao, Y.~Jiang, and Q.~Yin, ``Extensive facial landmark
  localization with coarse-to-fine convolutional network cascade,'' in
  \emph{Proceedings of the IEEE international conference on computer vision
  workshops}, 2013, pp. 386--391.

\bibitem{christossagonas_2013_55}
C.~Sagonas, G.~Tzimiropoulos, S.~Zafeiriou, and M.~Pantic, ``300 faces
  in-the-wild challenge: The first facial landmark localization challenge,'' in
  \emph{Proceedings of the IEEE international conference on computer vision
  workshops}, 2013, pp. 397--403.

\bibitem{kmesser_1999_54}
K.~Messer, J.~Matas, J.~Kittler, J.~Luettin, G.~Maitre \emph{et~al.},
  ``Xm2vtsdb: The extended m2vts database,'' in \emph{Second international
  conference on audio and video-based biometric person authentication}, vol.
  964.\hskip 1em plus 0.5em minus 0.4em\relax Citeseer, 1999, pp. 965--966.

\bibitem{martinkostinger_2011_50}
M.~Koestinger, P.~Wohlhart, P.~M. Roth, and H.~Bischof, ``Annotated facial
  landmarks in the wild: A large-scale, real-world database for facial landmark
  localization,'' in \emph{2011 IEEE international conference on computer
  vision workshops (ICCV workshops)}.\hskip 1em plus 0.5em minus 0.4em\relax
  IEEE, 2011, pp. 2144--2151.

\bibitem{Zhao_2022}
K.~Zhao, C.~Lin, K.~Liao, S.~Yang, and Y.~Zhao,
  ``\BIBforeignlanguage{en}{Revisiting radial distortion rectification in
  polar-coordinates: A new and efficient learning perspective},''
  \emph{\BIBforeignlanguage{en}{IEEE Transactions on Circuits and Systems for
  Video Technology}}, vol.~32, no.~6, p. 3552–3560, 2022.

\bibitem{diederikpkingma_2014_53}
D.~P. Kingma and J.~Ba, ``Adam: A method for stochastic optimization,'' in
  \emph{Proceedings of the International Conference on Learning Representations
  (ICLR)}, 2015, pp. 1--13.

\bibitem{jiankangdeng_2019_51}
J.~Deng, J.~Guo, E.~Ververas, I.~Kotsia, and S.~Zafeiriou, ``Retinaface:
  Single-shot multi-level face localisation in the wild,'' in \emph{Proceedings
  of the IEEE/CVF Conference on Computer Vision and Pattern Recognition
  (CVPR)}, 2020, pp. 5203--5212.

\bibitem{Yang_2021}
S.~Yang, C.~Lin, K.~Liao, C.~Zhang, and Y.~Zhao, ``Progressively complementary
  network for fisheye image rectification using appearance flow,'' in
  \emph{Proceedings of the IEEE/CVF Conference on Computer Vision and Pattern
  Recognition (CVPR)}, 2021, pp. 6348--6357.

\bibitem{sankhasmukherjee_2015_52}
S.~S. Mukherjee and N.~M. Robertson, ``Deep head pose: Gaze-direction
  estimation in multimodal video,'' \emph{IEEE Transactions on Multimedia},
  vol.~17, no.~11, pp. 2094--2107, 2015.

\bibitem{scaramuzza2006toolbox}
D.~Scaramuzza, A.~Martinelli, and R.~Siegwart, ``A toolbox for easily
  calibrating omnidirectional cameras,'' in \emph{2006 IEEE/RSJ International
  Conference on Intelligent Robots and Systems}.\hskip 1em plus 0.5em minus
  0.4em\relax IEEE, 2006, pp. 5695--5701.

\bibitem{guo2020towards}
J.~Guo, X.~Zhu, Y.~Yang, F.~Yang, Z.~Lei, and S.~Z. Li, ``Towards fast,
  accurate and stable 3d dense face alignment,'' in \emph{European Conference
  on Computer Vision}.\hskip 1em plus 0.5em minus 0.4em\relax Springer, 2020,
  pp. 152--168.

\end{thebibliography}
\bibliographystyle{IEEEtran}
\vspace{-12 mm}
\begin{IEEEbiography}[{\includegraphics[width=1in,height=1.25in,clip,keepaspectratio]{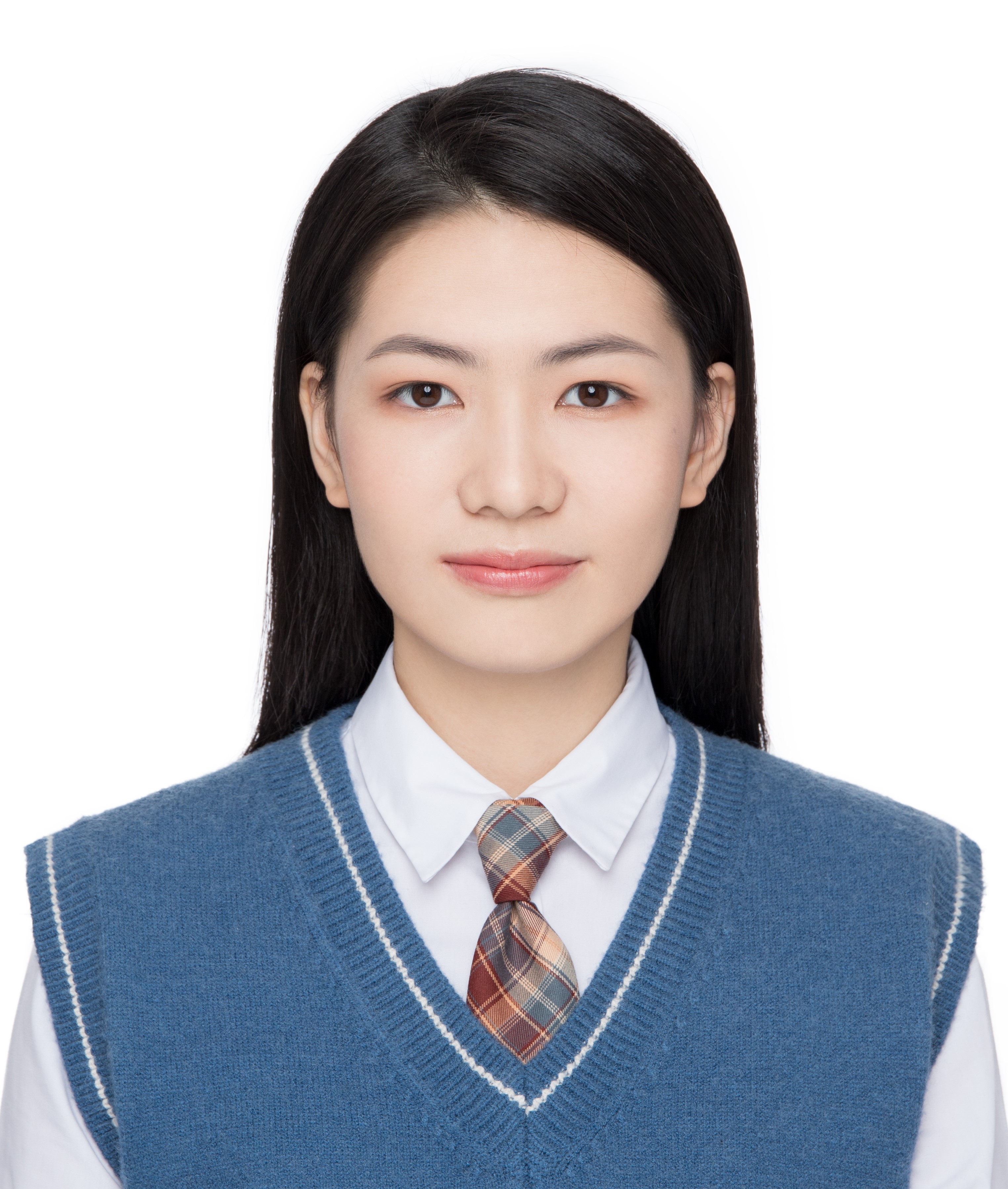}}]
{Bing Li} received her B.Eng. degree from Sun Yat-sen University, China, in 2021. She is currently a postgraduate student in the school of Electronics and Information Technology, Sun Yat-sen University. Her research interests include head pose estimation and computer vision. E-mail: libing29@mail2.sysu.edu.cn.
\end{IEEEbiography}
\vspace{-5 mm}

\begin{IEEEbiography}[{\includegraphics[width=1in,height=1.25in,clip,keepaspectratio]{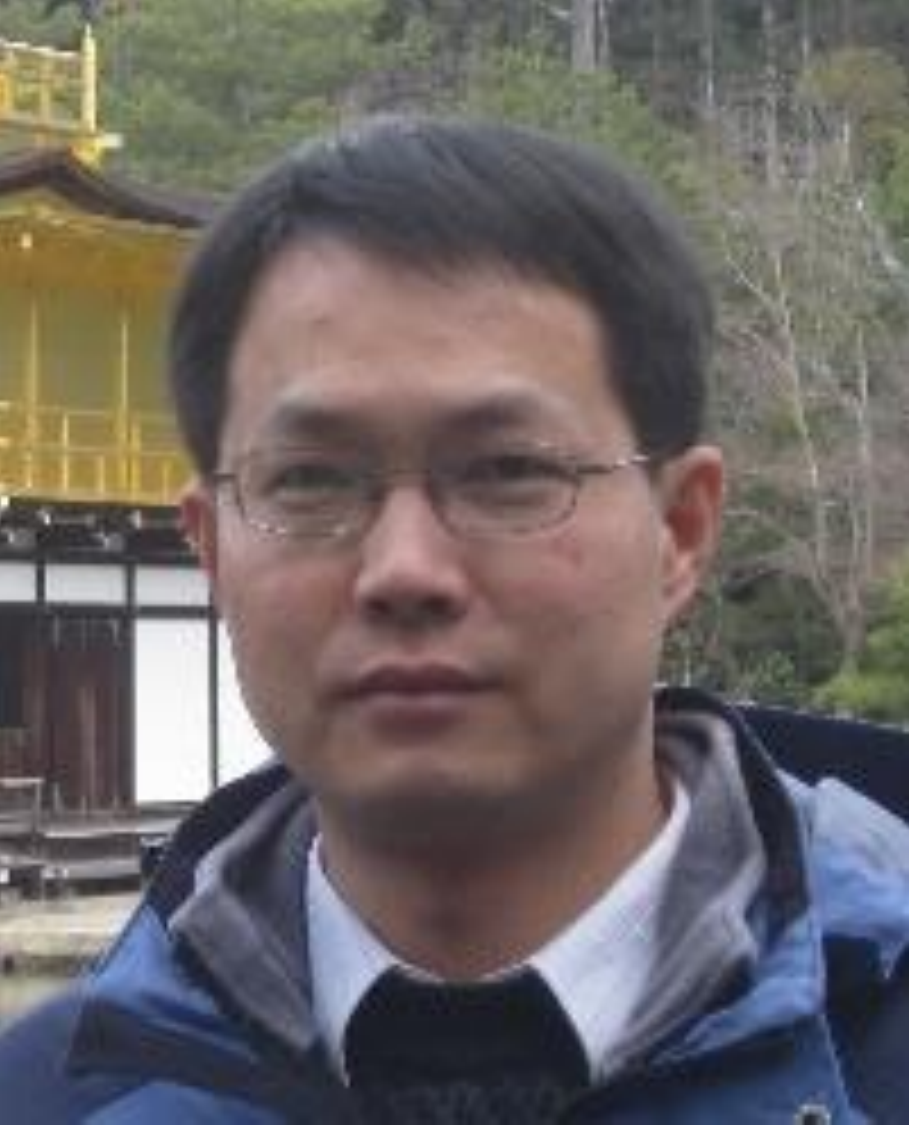}}]
{Dong Zhang} received his B.S.E.E. and M. S. degrees from Nanjing University, China, in 1999 and 2003, respectively, and Ph.D. degree from Sun Yat-sen University, China, in 2009. He is currently an associate professor in the school of Electronics and Information Technology, Sun Yat-sen University. His research interests include image processing, pattern recognition and information hiding.
\end{IEEEbiography}
\vspace{-5 mm}

\begin{IEEEbiography}[{\includegraphics[width=1in,height=1.25in,clip,keepaspectratio]{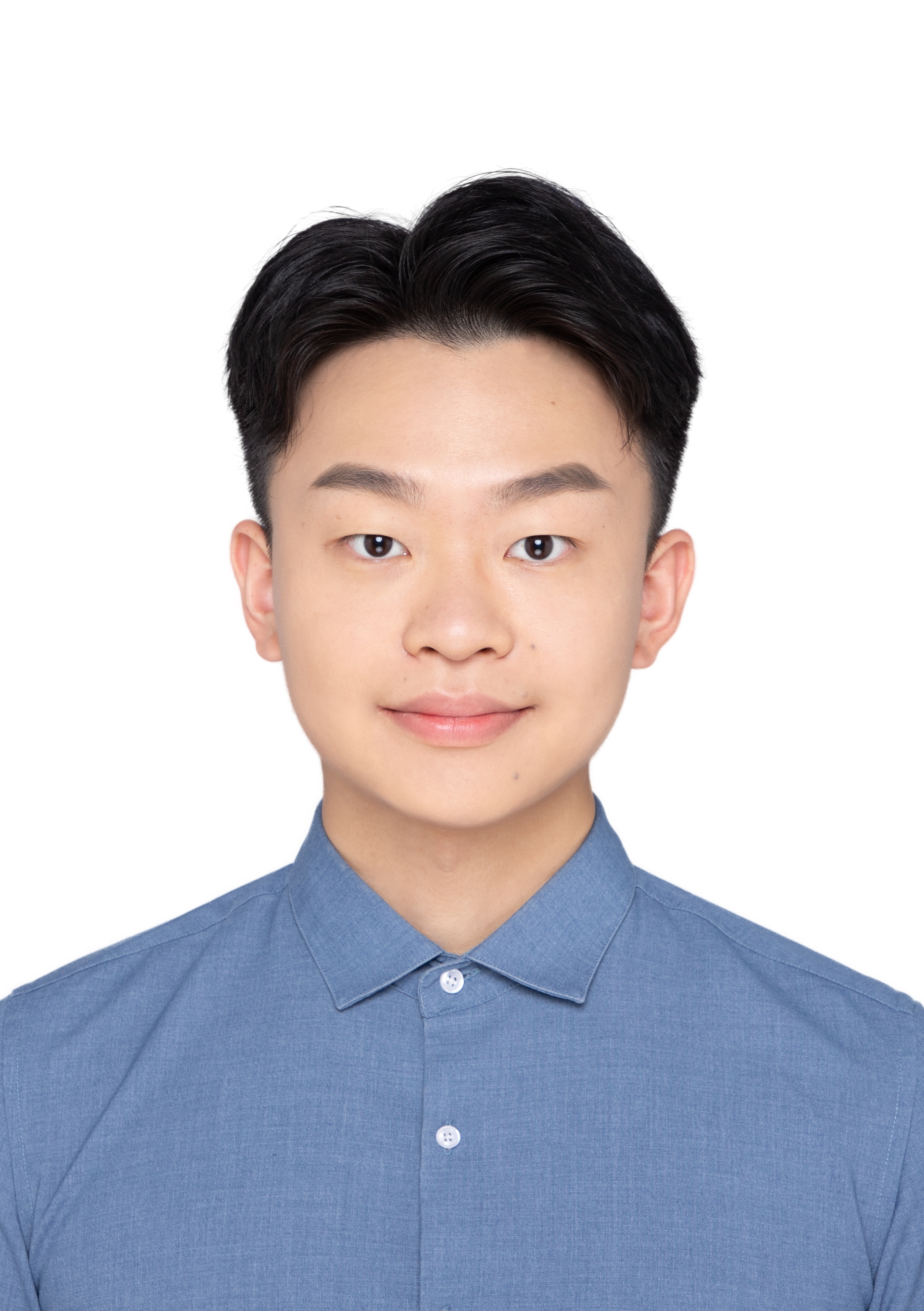}}]
{Cheng Huang} received his B.Eng. degree from Sun Yat-sen University, China, in 2021. He is currently a postgraduate student in the school of Electronics and Information Technology, Sun Yat-sen University. His research interests include group activity recognition and computer vision. E-mail: huangch63@mail2.sysu.edu.cn.
\end{IEEEbiography}
\vspace{-5 mm}

\begin{IEEEbiography}[{\includegraphics[width=1in,height=1.25in,clip,keepaspectratio]{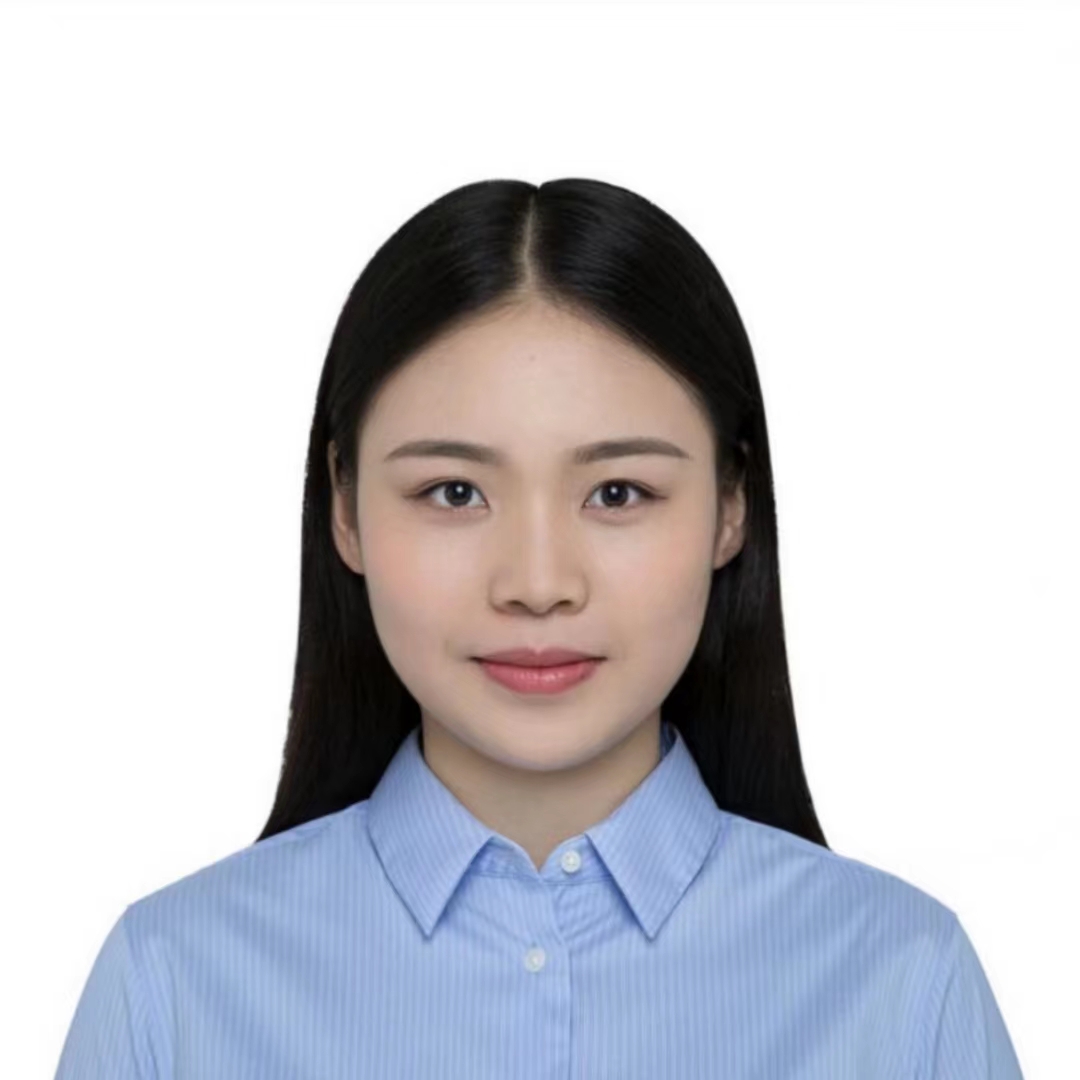}}]
{Yun Xian}  received the B.Eng. degree from the School of Information Science and Technology, Jinan University, Guangzhou, China, in 2021. She is currently pursuing the master's degree with the School of Information and Communication Engineering, Sun Yat-sen University. Her research interests include expression spotting and computer vision. E-mail: xiany7@mail2.sysu.edu.cn.
\end{IEEEbiography}
\vspace{-5 mm}

\begin{IEEEbiography}[{\includegraphics[width=1in,height=1.25in,clip,keepaspectratio]{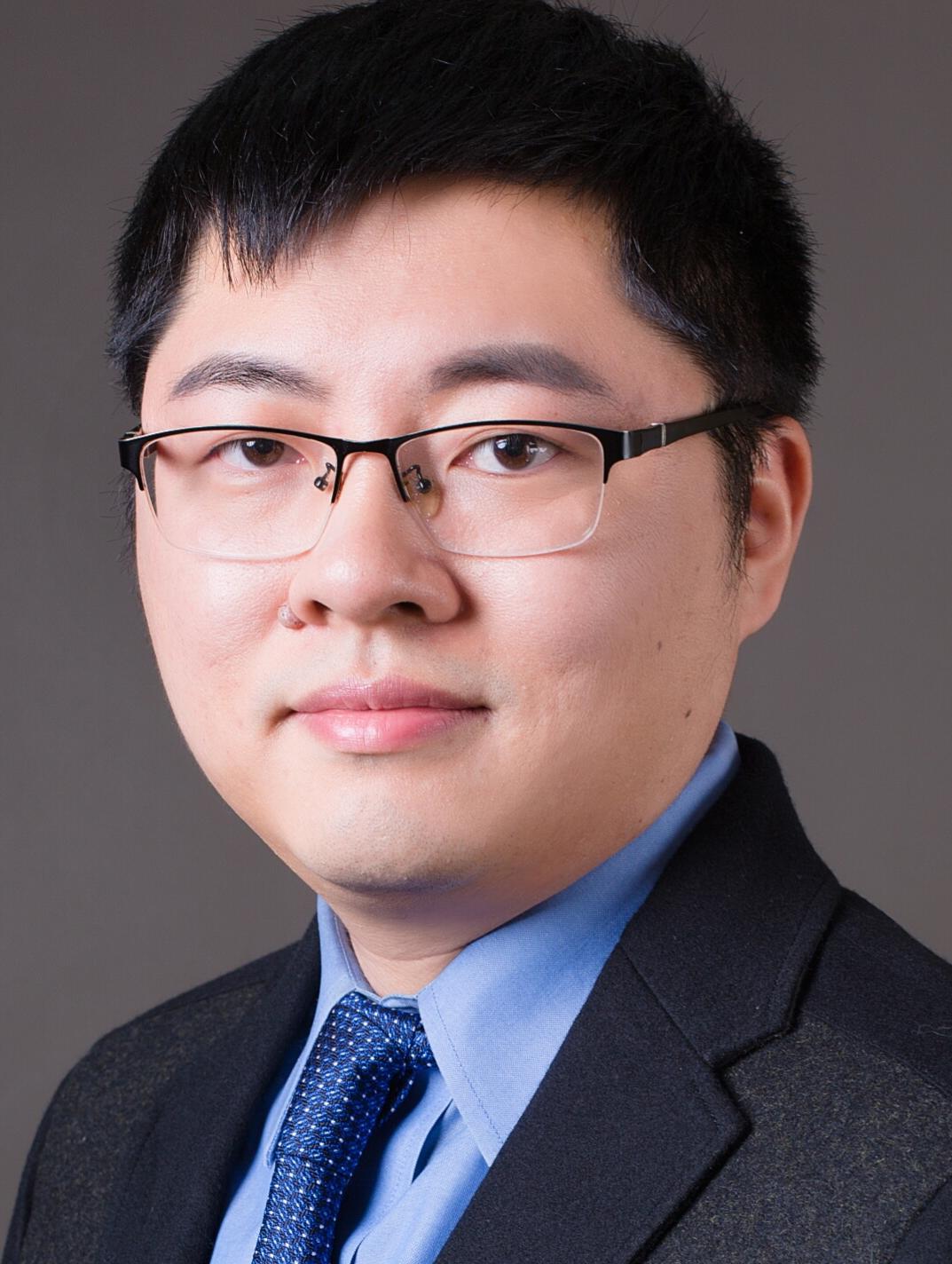}}]
{Ming Li} received his Ph.D. in Electrical Engineering from University of Southern California in May 2013. He is currently an associate professor of the Data Science Research Center at Duke Kunshan University, a research scholar at the ECE department of Duke University, and the adjunct professor at Wuhan University. His research interests are in the areas of speech processing and multimodal behavior signal analysis with applications to human centered behavioral informatics notably in health, education and security.
\end{IEEEbiography}
\vspace{-5 mm}

\begin{IEEEbiography}[{\includegraphics[width=1in,height=1.25in,clip,keepaspectratio]{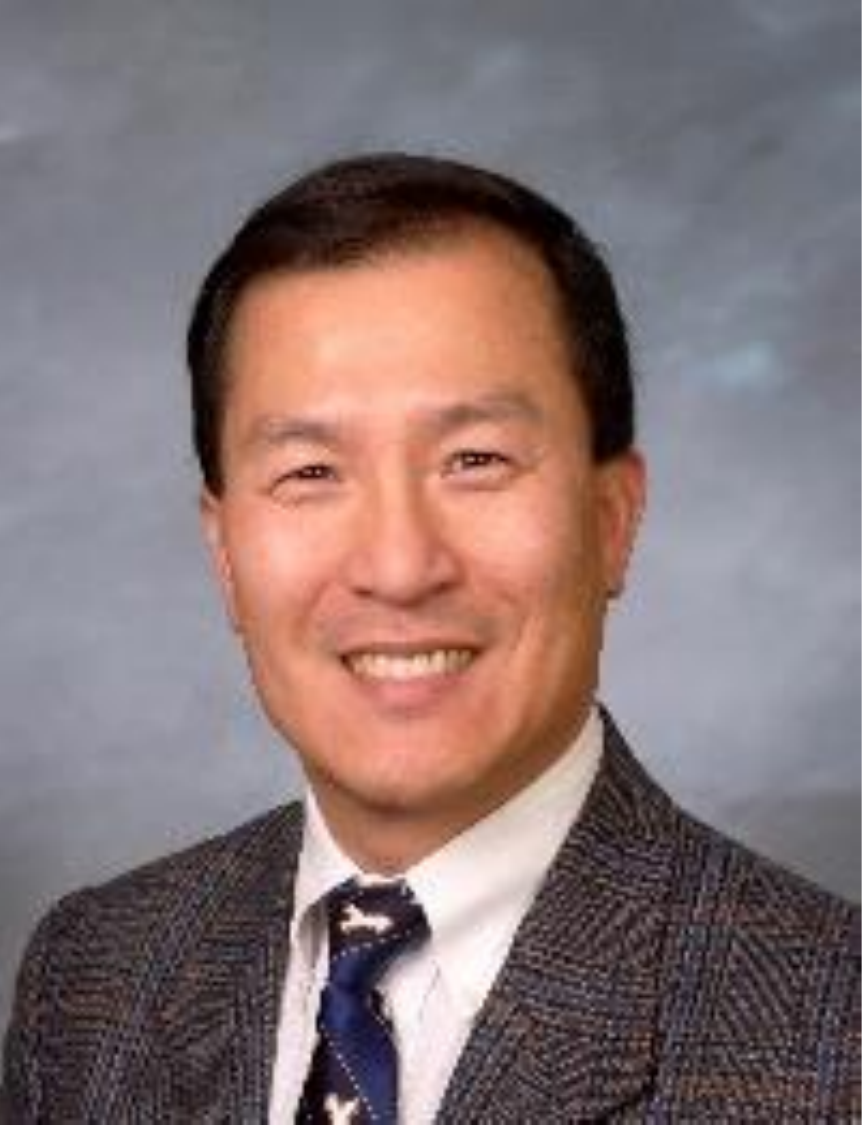}}]
{Dah-Jye Lee} received his M.S. and Ph.D. degrees in electrical engineering from Texas Tech University and MBA degree from Shenandoah University. He served in the machine vision industry for eleven years before joining Brigham Young University faculty in 2001. He is currently a professor and the director of the Robotic Vision Laboratory in the Electrical and Computer Engineering Department at BYU. His research focuses on artificial intelligence, high-performance visual computing, robotic vision, and visual inspection automation.
\end{IEEEbiography}		

\end{document}